\def\year{2021}\relax
\documentclass[letterpaper]{article} 
\usepackage{aaai21}  
\usepackage{times}  
\usepackage{helvet} 
\usepackage{courier}  
\usepackage[hyphens]{url}  
\usepackage{graphicx} 
\usepackage{graphicx}  
\usepackage{natbib}  
\usepackage{caption} 
\newcommand\round[1]{\left[#1\right]}

\title{Sparsity Aware Normalization for GANs}
\author{Idan Kligvasser, Tomer Michaeli \\}
\affiliations{
Technion–Israel Institute of Technology, Haifa, Israel \\
\{kligvasser@campus, tomer.m@ee\}.technion.ac.il \\
}

\newtheorem{example}{Example}
\newtheorem{lemma}{Lemma}

\usepackage{amsmath,amssymb}
\usepackage{color}
\usepackage[title]{appendix}
\usepackage{pifont}
\usepackage{algorithm}
\usepackage[noend]{algpseudocode}

\usepackage{multirow}
\usepackage{array}
\newcolumntype{L}[1]{>{\raggedright\let\newline\\\arraybackslash\hspace{0pt}}m{#1}}
\newcolumntype{C}[1]{>{\centering\let\newline\\\arraybackslash\hspace{0pt}}m{#1}}
\newcolumntype{R}[1]{>{\raggedleft\let\newline\\\arraybackslash\hspace{0pt}}m{#1}}

\DeclareMathOperator{\E}{\mathbb{E}}

\begin{document}

\maketitle

\begin{figure*}
\centering
\includegraphics[width=2.0\columnwidth]{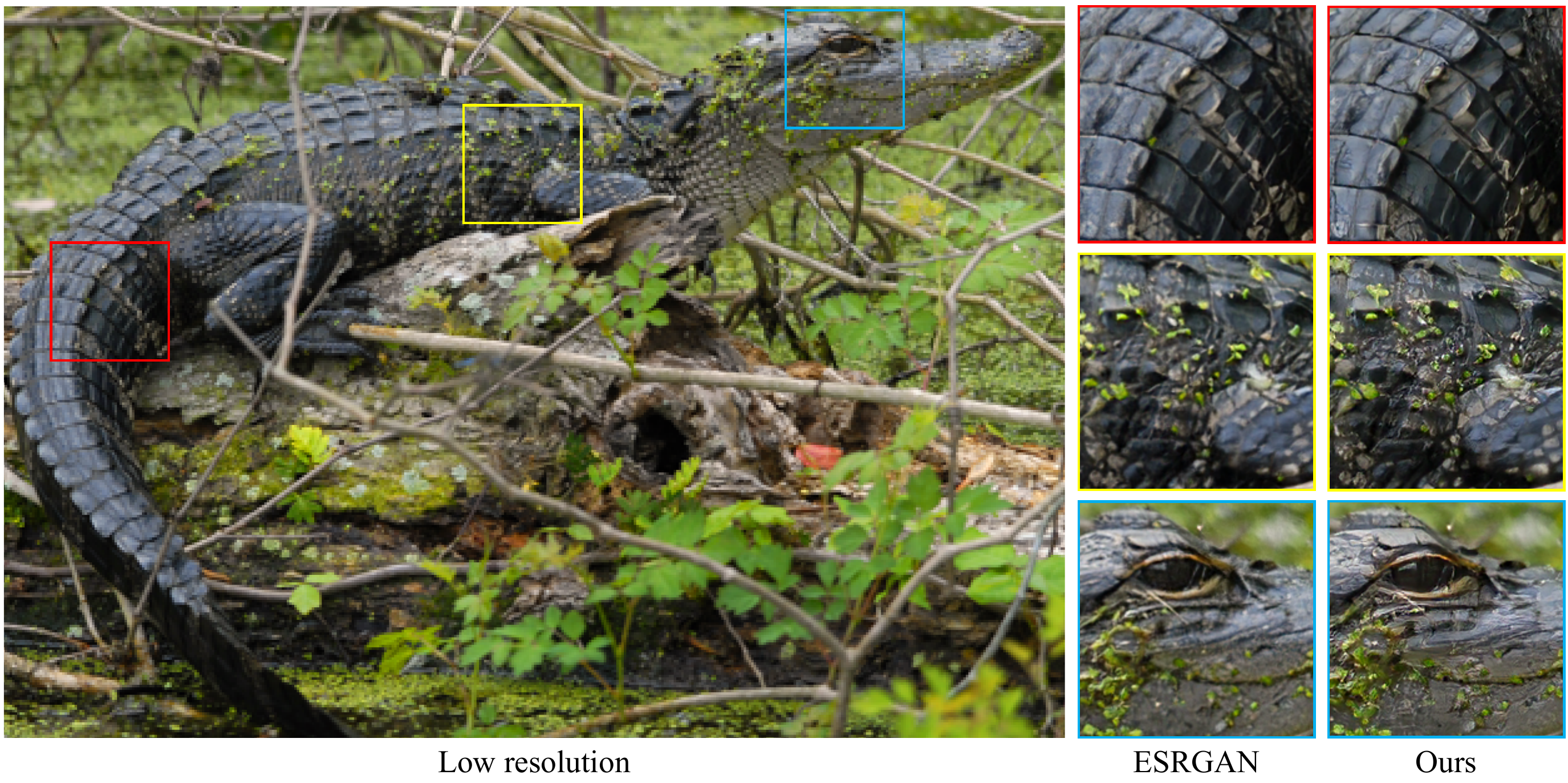}
\caption{\textbf{Super resolution with our sparsity aware normalization.} Our technique can boost the performance of any GAN-based method, while allowing less training epochs and smaller models. For example, in the task of $4\times$ super-resolution, we achieve more photo-realistic reconstructions than the state-of-the-art ESRGAN network \cite{esrgan}, while using a model with only $9\%$ the number of parameters of ESRGAN ($1.5$M for ours and $16.7$M for ESRGAN).}
\label{fig:sr}
\end{figure*}

\begin{abstract}
Generative adversarial networks (GANs) are known to benefit from regularization or normalization of their critic (discriminator) network during training. In this paper, we analyze the popular spectral normalization scheme, find a significant drawback and introduce  \emph{sparsity aware normalization} (SAN), a new alternative approach for stabilizing GAN training. As opposed to other normalization methods, our approach explicitly accounts for the sparse nature of the feature maps in convolutional networks with ReLU activations. We illustrate the effectiveness of our method through extensive experiments with a variety of network architectures. As we show, sparsity is particularly dominant in critics used for image-to-image translation settings. In these cases our approach improves upon existing methods, in less training epochs and with smaller capacity networks, while requiring practically no computational overhead.
\end{abstract}

\section{Introduction}
Generative adversarial networks (GANs) \cite{goodfellow2014generative} have made a dramatic impact on low-level vision and graphics, particularly in tasks relating to image generation \cite{radford2015unsupervised,karras2017progressive}, image-to-image translation \cite{isola2017image,zhu2017unpaired,stargan}, and single image super resolution \cite{srgan,esrgan,bahat2019explorable}. GANs can generate photo-realistic samples of fantastic quality \cite{karras2019style,brock2018large,shaham2019singan,srgan}, however they are often hard to train and require careful use of regularization and/or normalization methods for making the training stable and effective.

A factor of key importance in GAN training, is the way by which the critic (discriminator) network is optimized. An overly-sharp discrimination function can lead to gradient vanishing when updating the generator, while an overly-smooth function can lead to poor discrimination between real and fake samples and thus to insufficient supervision for the generator. One of the most successful training approaches, is that arising from the Wasserstein GAN (WGAN) \cite{arjovsky2017wasserstein} formulation, which asserts that the critic should be chosen among the set of Lipschitz-1 functions. Precisely enforcing this constraint is impractical \cite{NIPS2018_7640}, yet simple approximations, like weight clipping \cite{arjovsky2017wasserstein} and gradient norm penalty  \cite{gulrajani2017improved}, are already quite effective.

Perhaps the most effective approximation strategy is \emph{spectral normalization} \cite{spectral}. This method normalizes the weights of the critic network after every update step, in an attempt to make each layer Lipschitz-1 individually (which would guarantee that the end-to-end function is Lipschitz-1 as well). Due to its simplicity and its significantly improved results, this approach has become the method of choice in numerous GAN based algorithms (\emph{e.g.} \cite{miyato2018cgans,spade,brock2018large,armanious2020medgan}).

In this paper, we present a new weight normalization strategy that outperforms spectral normalization, as well as all other methods, by a significant margin on many tasks and with various network architectures (see \emph{e.g.}, Fig.~\ref{fig:sr}). We start by showing, both theoretically and empirically, that normalizing each layer to be Lipschitz-1 is overly restrictive. In fact, as we illustrate, such a normalization leads to very poor GAN training if done correctly. We identify that the real reason for the success of \cite{spectral} is actually its systematic bias in the estimation of the Lipschitz constant for convolution layers, which is typically off by roughly a factor of 4. Following our analysis, we show that a better way to control the end-to-end smoothness of the critic, is to normalize each layer by its amplification of the \emph{typical signals} that enter it (rather than the worst-case ones). As we demonstrate, in convolutional networks with ReLU activations, these signals are typically channel-sparse (namely many of their channels are identically zero). This motivates us to suggest \emph{sparsity aware normalization} (SAN).

Our normalization has several advantages over spectral normalization. First, it leads to better visual results, as also supported by quantitative evaluations with the Inception score (IS) \cite{inception} and the Fr\'echet Inception distance (FID) \cite{fid}. This is true in both unconditional image generation and conditional tasks, such as label-to-image translation, super-resolution, and attribute transfer. Second, our approach better stabilizes the training, and it does so at practically no computational overhead. In particular, even if we apply only a single update step of the critic for each update of the generator, and normalize its weights  only once every 1K steps, we still obtain an improvement over spectral normalization. Finally, while spectral normalization benefits from different tuning of the optimization hyper-parameters for different tasks, our approach works well with the precise same settings for all tasks.

\section{Rethinking Per-Layer Normalization}
GANs \cite{goodfellow2014generative} minimize the distance between the distribution of their generated ``fake'' samples, $\mathbb{P}_{\rm F}$, and the distribution of real images, $\mathbb{P}_{\rm R}$, by diminishing the ability to discriminate between samples drawn from $\mathbb{P}_{\rm F}$ and samples drawn from~$\mathbb{P}_{\rm R}$. In particular, the Wasserstein GAN (WGAN)~\cite{arjovsky2017wasserstein} targets the minimization of the Wasserstein distance between $\mathbb{P}_{\rm F}$ and $\mathbb{P}_{\rm R}$, which can be expressed as
\begin{equation}
W(\mathbb{P}_{\rm R},\mathbb{P}_{\rm F})=\sup_{\|f\|_L\leq 1} \mathbb{E}_{x\sim \mathbb{P}_{\rm R}}[f(x)] - \mathbb{E}_{x\sim \mathbb{P}_{\rm F}}[f(x)].
\end{equation}
Here, the optimization is over all critic functions $f:\mathbb{R}^n\rightarrow\mathbb{R}$ whose Lipschitz constant is no larger than $1$. Thus, the critic's goal is to output large values for samples from $\mathbb{P}_{\rm R}$ and small values for samples from $\mathbb{P}_{\rm F}$. The GAN's generator attempts to shape the distribution of fake samples, $\mathbb{P}_{\rm F}$, so as to minimize $W(\mathbb{P}_{\rm R},\mathbb{P}_{\rm F})$ and so to rather decrease this gap.

The Lipschitz constraint has an important role in the training of WGANs, as it prevents overly sharp discrimination functions that hinder the ability to update the generator. However, since $f$ is a neural network, this constraint is impractical to enforce precisely \cite{NIPS2018_7640}, and existing methods resort to rather inaccurate approximations. Perhaps the simplest approach is to clip the weights of the critic network \cite{arjovsky2017wasserstein}. However, this leads to stability issues if the clipping value is taken to be too small or too large. An alternative, is to penalize the norm of the gradient of the critic network \cite{gulrajani2017improved}. Yet, this often has poor generalization to points outside the support of the current generative distribution.

To mitigate these problems, Miyato~\emph{et~al.}~(\citeyear{spectral}) suggested to enforce the Lipschitz constraint on each layer individually. Specifically, denoting the function applied by the $i$th layer by $\phi_i(\cdot)$, we can write
\begin{equation}
f(x)=(\phi _N \circ \phi _{N-1} \circ .. \circ \phi _1)(x). 
\label{eq:fucntion_composition}
\end{equation}
Now, since
$\|\phi _1 \circ \phi_2\|_L \leq \|\phi _1 \|_L \cdot \|\phi _2 \|_L$,
we have that
\begin{equation}
\|f\|_L \leq \|\phi _N\|_L \cdot \|\phi _{N-1}\|_L \cdot ... \cdot \|\phi _1 \|_L.
\label{eq:lipschitz_layerwise}
\end{equation}
This implies that restricting each $\phi_i$ to be Lipschitz-1, ensures that $f$ is also Lipschitz-1. Popular activation functions, such as ReLU and leaky ReLU, are Lipschitz-1 by construction. For linear layers (like convolutions), ensuring the Lipschitz condition merely requires normalizing the weights by the Lipschitz constant of the transform, which is the top singular value of the corresponding weight matrix. 

\begin{figure}[t]
\centering
\includegraphics[width=0.9\columnwidth]{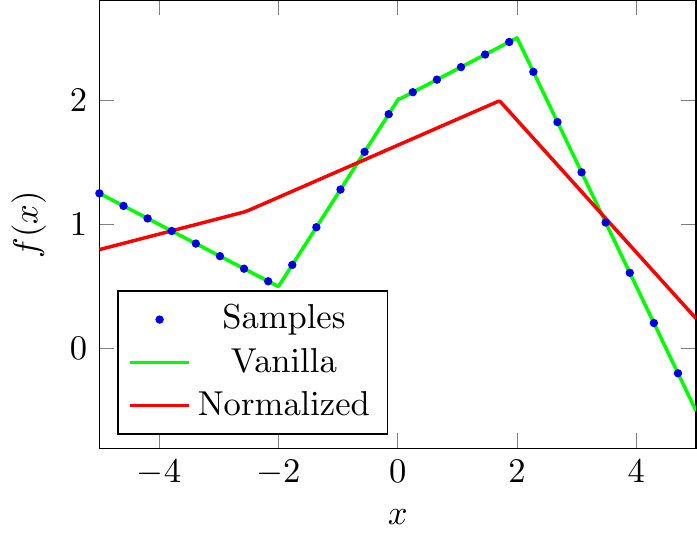}
\caption{\textbf{Fitting to a Lipschitz-1 function.} Here, we trained a network with one hidden layer to fit to samples of a Lipschitz-1 function (blue dots). When using vanilla training without any normalization, the fit is perfect (green). However, when using layer-wise spectral normalization, the fit is poor (red). This illustrates the fact that the set of functions that can be represented by a network with Lipschitz-1 layers is often significantly smaller than the set of all Lipschitz-1 functions that can be represented by the same architecture.} 
\label{fig:example1}
\end{figure}	

\begin{figure}[t]
\centering
\includegraphics[width=0.9\columnwidth]{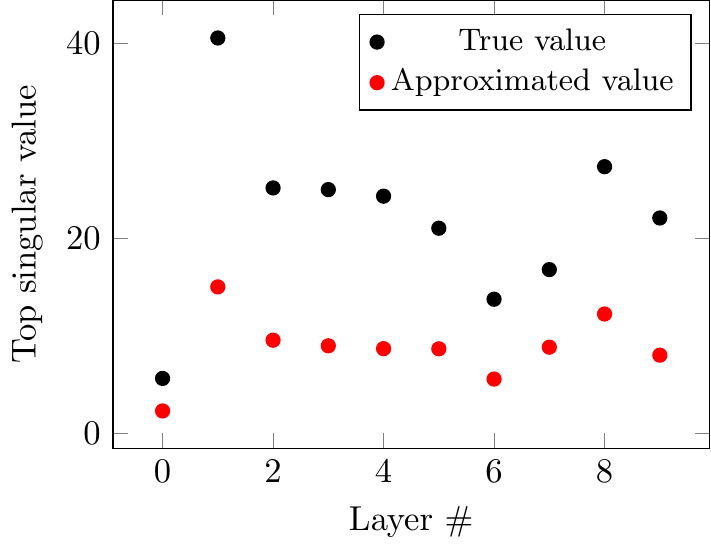}
\caption{\textbf{Top singular value.} We plot the top singular value (black) of each convolution layer of a trained ResNet critic network, as well as its approximation employed by~\cite{spectral} (red). The approximation is typically much smaller than the actual value, implying that the weights after normalization are in fact much larger than intended.} 
\label{fig:top_singular}
\end{figure}	

This per-layer normalization strategy has gained significant popularity due to its simplicity and the improved results that it provides when compared to the preceding alternatives. However, close inspection reveals that normalizing each layer by its top singular value is actually too conservative. That is, restricting each layer to be Lipschitz-1, typically leads to a \emph{much smaller set of permissible functions} than the set of functions whose end-to-end Lipschitz constant is 1. As a simple illustration, consider the following example (see Appendix A).
\begin{example}
	Let $f:\mathbb{R}\rightarrow\mathbb{R}$ be a two-layer network with
	\begin{equation}
	\phi_1(x)= \sigma(w_1 x + b_1),\qquad \phi_2(z)= w_2^T z + b_2,  
	\end{equation}
 where $\sigma$ is the ReLU activation function, $w_1,w_2,b_1\in\mathbb{R}^n$, and $b_2\in\mathbb{R}$. 
	Such a critic can implement any continuous piece-wise linear function with $n+1$ segments. Now, the end-to-end constraint $\|f\|_L\leq 1$, restricts the slope of each segment to satisfy $|f'(x)|\leq 1$. But the layer-wise constraints\footnote{Since $w_1$ and $w_2$ are $n\times 1$, their top singular value is simply their Euclidean norm.} $\|w_1\|\leq 1$, $\|w_2\|\leq 1$, allow a much smaller set of functions, as they also impose for example that $|f'(-\infty)+f'(\infty)|\leq 1$. In particular, they rule out the identity function $f(x)=x$, and also any function with slope larger than $0.5$ or smaller than $-0.5$ simultaneously for $x\rightarrow\infty$ and $x\rightarrow-\infty$. This is illustrated in Fig.~\ref{fig:example1}.
\end{example}

This example highlights an important point. When we normalize a layer by its top singular value, we restrict how much it can amplify an arbitrary input. However, this is overly pessimistic since not all inputs to that layer are admissible. In the example above, for most choices of $w_1$ the input to the second layer is necessarily sparse because of the ReLU. Specifically, if $b_1=0$ and $w_1$ has $k_p$ positive entries and $k_n$ negative ones, then the output of the first layer cannot contain more than $\max\{k_p,k_n\}$ non-zero entries. This suggests that when normalizing the second layer, we should only consider how much it amplifies sparse vectors.

As a network gets deeper, the attenuation caused by such layer-wise normalization accumulates, and severely impairs the network's representation power. One may wonder, then, why the layer-wise spectral normalization of \cite{spectral} works in practice after all. The answer is that for convolutional layers, this method uses a very crude approximation of the top singular value, which is typically $4\times$ smaller than the true top singular value\footnote{In~\cite{spectral}, the top singular value of the convolution operation is approximated by the top singular value of a 2D matrix obtained by reshaping the 4D kernel tensor.}. We empirically illustrate this in Fig.~\ref{fig:top_singular} for a ResNet critic architecture, where we use the Fourier domain formulation of \cite{sedghi2018singular} to compute the true top singular value. This observation implies that in \cite{spectral}, the weights after normalization are in fact much larger than intended.

\begin{figure}[t]
\centering
\includegraphics[width=1.0\columnwidth]{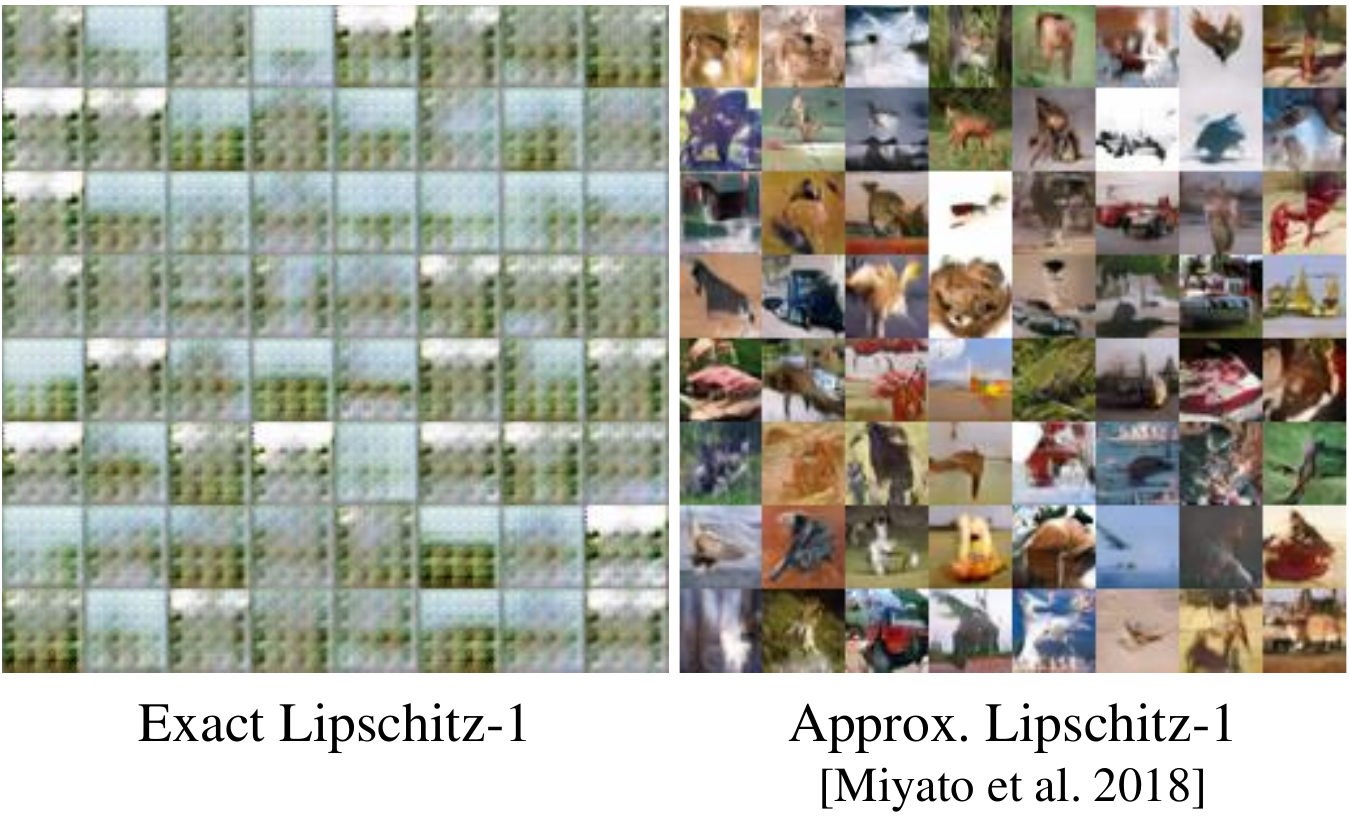}
\caption{\textbf{The effect of normalization}. Here, we trained a WGAN on the CIFAR-10 dataset using exact per-layer spectral normalization and the approximate method of \cite{spectral}, both initialized with the same draw of random weights. The exact normalization does not converge, while the approximate one rather leads to good conditioning.
}\label{fig:GenerationNormMethods}
\end{figure}

What would happen had we normalized each layer by its true top singular value? As shown in Fig.~\ref{fig:GenerationNormMethods}, in this case, the training completely fails. This is because the weights become extremely small and the gradients vanish.

\section{Sparsity Aware Normalization}
We saw that the spectral normalization of \cite{spectral} is effective because of the particular approximation used for $\|\phi_i\|_L$. A natural question, then, is whether we can somehow improve upon this normalization scheme. A naive approach would be to set a multiplier parameter $\sigma$, to adjust their normalization constant. However, as the authors of \cite{spectral} themselves indicate, such a parameter does not improve their results. This implies that the set of discriminator functions satisfying their per-layer constraints does not overlap well with the set of Lipschitz-1 functions as neither dilation nor erosion of this set improves their results.

\begin{figure}[t]
\centering
\includegraphics[width=0.9\columnwidth]{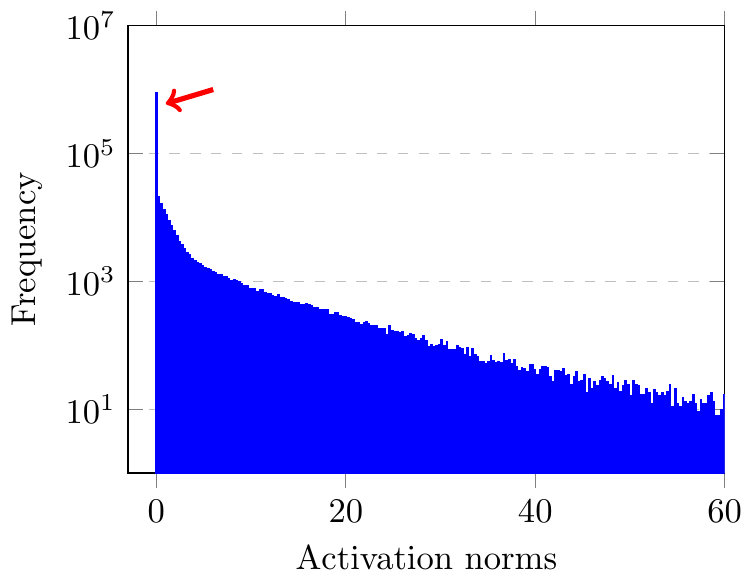}
\caption{\textbf{Channel sparsity.} The histogram of the channel norms at the last layer of a critic trained without normalization. For most input images, many of the channels are identically zero.} 
\label{fig:sparse_vanilla}
\end{figure}

A more appropriate strategy is therefore to seek for a normalization method that explicitly accounts for the statistics of signals that enter each layer. An important observation in this respect, is that in convolutional networks with ReLU activations, the features are typically channel-sparse. That is, for most input signals, many of the channels are identically zero. This is illustrated in Fig.~\ref{fig:sparse_vanilla}, which shows a histogram of the norms of the channels of the last layer of a trained critic\footnote{We used no normalization, but chose a run that converged.}, computed over $2048$ randomly sampled images from the training set.

In light of this observation, rather than normalizing a layer $\phi(x)=Wx+b$ by its Lipschitz constant,
\begin{equation}
\|\phi\|_L = \sup_{\|x\| \leq 1} \|Wx\|,
\label{eq:lipschitz_linear}
\end{equation}
here we propose to modify the constraint set to take into account only channel-sparse signals. Moreover, since we know that many output channels are going to be zeroed out by the ReLU that follows, we also modify the objective of~\eqref{eq:lipschitz_linear} to consider the norm of each output channel individually.

Concretely, for a multi-channel signal $\boldsymbol{x}$ with channels $x_1,\ldots,x_k$, let us denote by $\|\boldsymbol{x}\|_\infty$ its largest channel norm, $\max\{\|x_i\|\}$, and by $\|\boldsymbol{x}\|_0$ its number of nonzero channels, $\#\{\|x_i\|>0\}$. 
With these definitions, we take our normalization constant to be\footnote{Note that $\|\cdot\|_{0,\infty}$ is not a norm since $\ell_0$ is not a norm.}
\begin{equation}
\|W\|_{0,\infty} \triangleq \sup_{\|\boldsymbol{x}\|_0 \leq 1\atop\|\boldsymbol{x}\|_\infty \leq 1} \|W\boldsymbol{x}\|_\infty.
\label{eq:SAN_linear}
\end{equation}
Normalizing by $\|W\|_{0,\infty}$ ensures that there exists no $1$-sparse input signal (\emph{i.e.}~with a single nonzero channel) that can cause the norm of some output channel to exceed $1$.

For convolutional layers, computing $\|W\|_{0,\infty}$ is simple. Specifically, if $W$ has $n$ input channels and $m$ output channels, then the $i$th channel of $\boldsymbol{y}=W\boldsymbol{x}$ can be expressed as
\begin{equation}
y_i = \sum_{j=1}^n w_{i,j}*x_j.
\end{equation}
Here, `$*$' denotes single-input-single-output convolution and $w_{i,j}$ is the kernel that links input channel $j$ with output channel $i$. Now, using the kernels $\{w_{i,j}\}$, we can compute $\|W\|_{0,\infty}$ as follows  (see Appendix B).

\begin{lemma}\label{lem:SparseNorm}
	For a multiple-input-multiple-output filter $W$ with cyclic padding, 
\begin{equation}
\|W\|_{0,\infty} = \max_{i,j}\|\mathcal{F}\{w_{i,j}\}\|_\infty,
\label{eq:lemma}
\end{equation}
where $\mathcal{F}\{w_{i,j}\}$ is the discrete Fourier transform of $w_{i,j}$, zero-padded to the spatial dimensions of the channels.
\end{lemma}

Thus, to compute our normalization constant, all we need to do is take the Fourier transform of each kernel, find the maximal absolute value in the transform domain, and then take the largest among these $m\times n$ top Fourier values.

\subsection{Efficiency}

To take advantage of Lemma~\ref{lem:SparseNorm}, we use cyclic padding for all convolutional layers of the critic. This allows us to employ the fast Fourier transform (FFT) for computing the normalization constants of the layers. For  fully-connected layers, we use the top singular value of the eight matrix, as in \cite{spectral}. The overhead in running time is negligible. For example, on CIFAR-10, each critic update takes the same time as spectral normalization and $20\%$ less than gradient-penalty regularization (see Appendix F).

In models for large images, storing the FFTs of all the filters of a layer can be prohibitive. In such settings, we compute the maximum in \eqref{eq:lemma} only over a random subset of the filters. We compensate for our under-estimation of the maximum by multiplying the resulting value by a scalar~$g$. As we show in the Appendix, the optimal value of $g$ varies very slowly as a function of the percentage of chosen filters (\emph{e.g.} it typically does not exceed $1.3$ even for ratios as low as $25\%$). This can be understood by regarding the kernels' top Fourier coefficients as independent draws from some density. When this density decays fast, the expected value of the maximum over $k$ draws increases very slowly for large $k$. For example, for the exponential distribution (which we find to be a good approximation), we show in the Appendix that the optimal $g$ for ratio $r$ is given by 
\begin{equation}
g=\frac{\sum_{j=1}^{m\cdot n} \frac{1}{m\cdot n-j+1}}{\sum_{j=1}^{\round{m \cdot n \cdot r}} \frac{1}{\round{m \cdot n \cdot r}-j+1}},
\label{eq:compensation_factor}
\end{equation}
leading to \emph{e.g.} $g\approx 1.2$ for $r=25\%$ with $m\cdot n=64^2$ filters.

\begin{figure}[t]
\centering
\includegraphics[width=0.9\columnwidth]{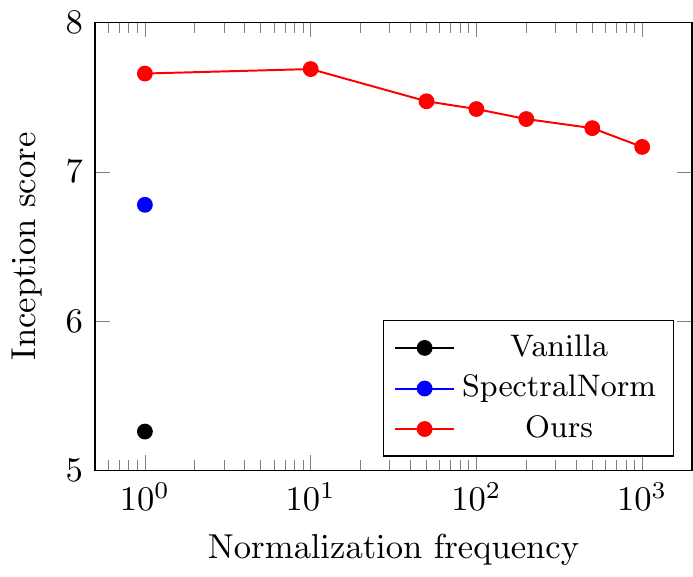}
\caption{\textbf{Efficiency.} Here we compare three WGANs, trained for 100 epochs on the CIFAR-10 dataset \cite{cifar}: (i) without normalization, (ii) with spectral-normalization \cite{spectral}, (iii) with our normalization. The training configurations and the initial seed are the same for all networks. In contrast to \cite{spectral}, which performs weight normalization after each critic update, in our method we can normalize the layers much less frequently. Pay attention that even if we normalize only once every $1000$ steps, less than $80$ updates in total, we still outperform spectral normalization by a large margin.} 
\label{fig:efficiency}
\end{figure}	

The effect of our normalization turns out to be very strong and therefore, besides a boost in performance, it also allows more efficient training than spectral normalization \cite{spectral} and other WGAN methods. Particularly:
	
	\noindent\textbf{Critic updates:} For every update step of the generator, we perform only one update step of the critic. This is in contrast to other WGAN schemes, which typically use at least three \cite{arjovsky2017wasserstein,gulrajani2017improved,spectral}. Despite the fewer updates, our method converges faster (see Appendix E). 
	
	\noindent\textbf{Normalization frequency:} In spectral normalization, the weights are normalized after each critic update (using a single iteration of the power method). In contrast, we can normalize the layers much less frequently and still obtain a boost in performance. For example, as shown in Fig.~\ref{fig:efficiency}, even if we normalize only once every 1000 steps, we still outperform spectral normalization by a large margin.
	
	\noindent\textbf{Hyper-parameters:} As opposed to other normalization methods, like \cite{arjovsky2017towards,gulrajani2017improved,ioffe2015batch,salimans2016weight,brock2016neural,spectral}, our algorithm does not require special hyper-parameter tuning for different tasks. All our experiments use the same hyper-parameters.

\section{Experiments}
We now demonstrate the effectiveness of our approach in several  tasks. In all our experiments, we apply normalization after each critic update step to obtain the best results.

\addtolength{\tabcolsep}{3pt} 
\begin{table}[t]
\begin{center}
\begin{tabular}{p{3cm} c c}
Method & CIFAR-10 & STL-10 \\ \hline \hline
Real-data & $11.24\pm .12$ & $26.08\pm .26$ \\ \hline
\textbf{Unconditional GAN (Standard CNN)} && \\
Weight clipping & $6.41\pm .11$ & $7.57 \pm .10$ \\
WGAN-GP & $ 6.68\pm .06$ & $ 8.42\pm .13$\\
Batch norm & $ 6.27 \pm .10$ &  \\
Layer norm & $ 7.19 \pm .12$ & $ 7.61\pm .12$\\
Weight norm & $6.84 \pm .07$ & $ 7.16\pm .10$\\
Orthonormal & $7.40 \pm .12$ & $8.67 \pm .08$\\
SN-GANs & $7.58 \pm .12$ & $8.79 \pm .14$\\
(ours) SAN-GANs & $\mathbf{7.89} \pm .09$ &  $\mathbf{9.18} \pm .06$\\ \hline
\textbf{Unconditional GAN (ResNet)} && \\
Orthonormal & $7.92 \pm .04$ & $8.72 \pm .06$ \\
SN-GANs & $8.22 \pm .05$ & $9.10 \pm .04$\\
(ours) SAN-GANs & $\mathbf{8.43} \pm .13$ & $\mathbf{9.21} \pm .10$\\  \hline
\textbf{Conditional GAN (ResNet)} &  & \\
BigGAN & $9.24 \pm .16$ & \\
(ours) SAN-BigGAN & $\mathbf{9.53} \pm .13$ & \\

\end{tabular}
\end{center}
\caption{Inception scores for image generation on the CIFAR-10 and STL-10 datasets. The SAN method outperforms all other regularization methods.}
\label{tab:generation}
\end{table}
\addtolength{\tabcolsep}{-3pt} 

\subsection{Image Generation}\label{image_generation}
We start by performing image generation experiments on the CIFAR-10 \cite{cifar} and STL-10 \cite{stl} datasets. We use these simple test-beds only for the purpose of comparing different regularization methods on the same architectures. Here, we use $r=100\%$ of the filters (and thus a compensation of $g=1$).

Our first set of architectures is that used in~\cite{spectral}. But to showcase the effectiveness of our method, in our STL-10 ResNet critic we remove the last residual block, which cuts its number of parameters by $75\%$, from $19.5$M to $4.8$M (the competing methods use the $19.5$M variant). The architectures are described in full in the Appendix. As in \cite{spectral}, we use the hinge loss \cite{wang2017magan} for the critic's updates. We train all networks for 200 epochs with batches of 64 using the Adam optimizer \cite{adam}. We use a learning rate of $2\cdot 10^{-4}$ and momentum parameters $\beta_1=0.5$ and $\beta_2=0.9$.

Additionally, we experiment with the more modern BigGAN architecture \cite{brock2018large} for conditional generation on CIFAR-10. We replace the spectral normalization by our SAN in all critic's res-blocks, and modify Adam's first momentum parameter to $\beta_1=0.5$.

\begin{figure}[t]
\centering
\includegraphics[width=0.9\columnwidth]{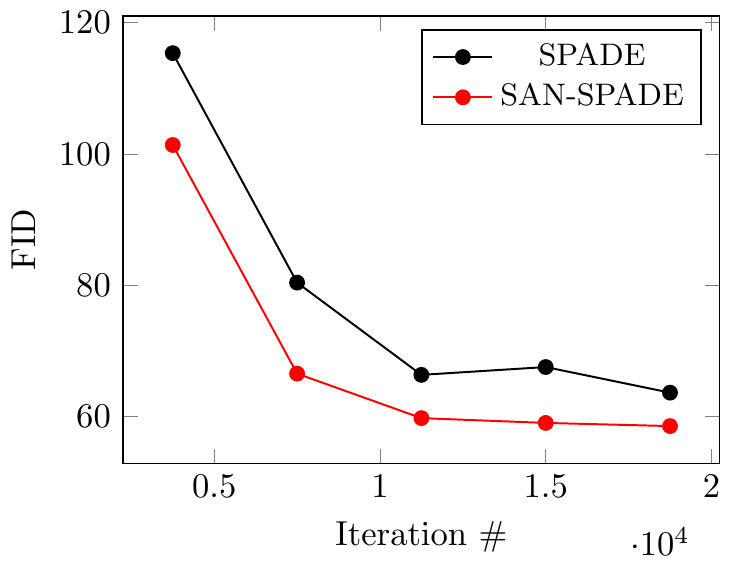}
\caption{\textbf{Model convergence in image translation.} The FID score of our SAN-SPADE converges faster and to a better result than the original SPADE.} 
\label{fig:fids}
\end{figure}

\begin{figure}[t]
\centering
\includegraphics[width=1.0\columnwidth]{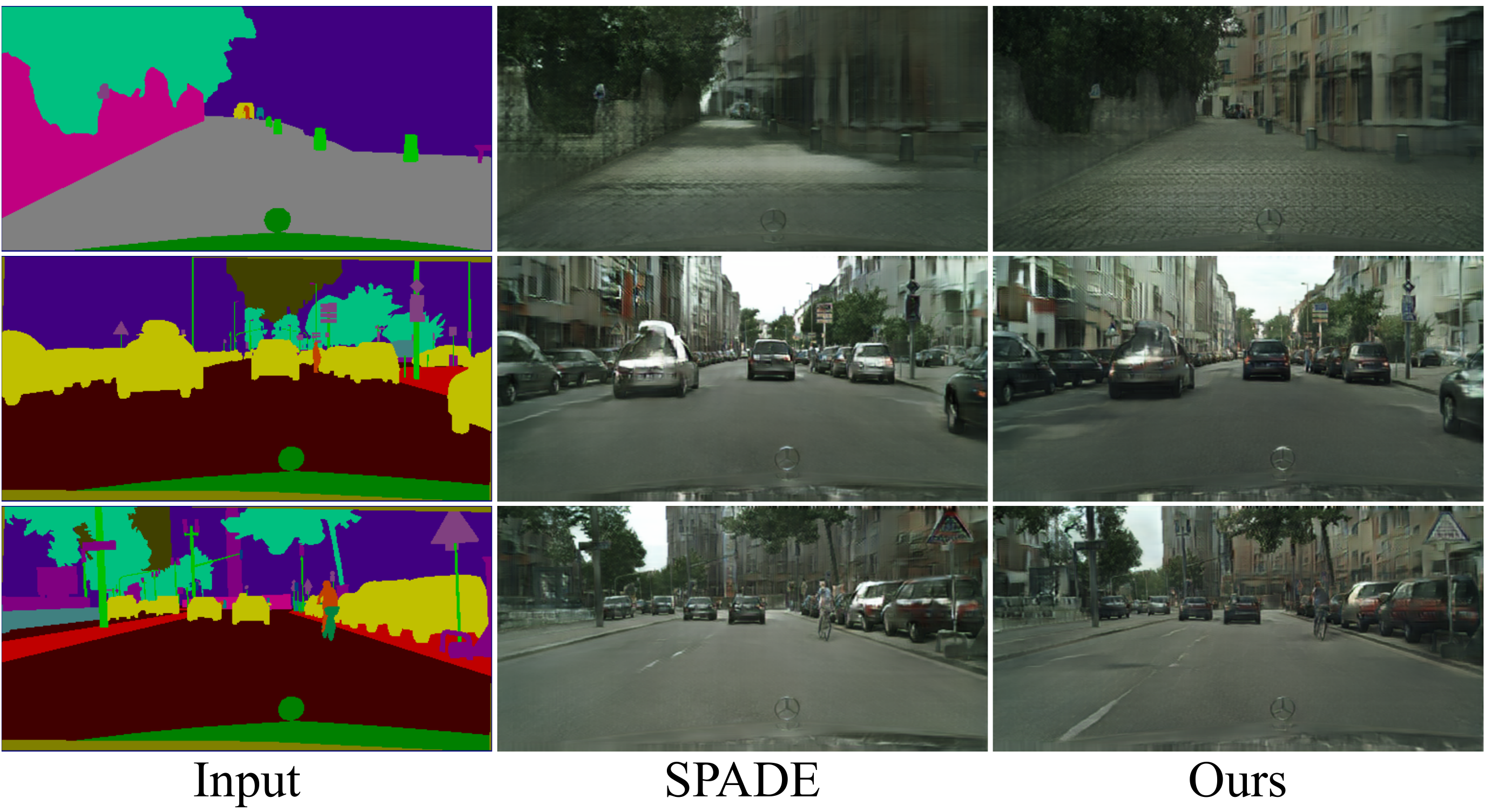}
\caption{\textbf{Visual comparison for image translation.} The images synthesized by our SAN-SPADE have less artifacts and contain more fine details than the original SPADE.} 
	\label{fig:image2image_spade}
\end{figure}

Table~\ref{tab:generation} shows comparisons between our approach (SAN-GAN) and other regularization methods in terms of Inception score \cite{inception}. The competing methods include weight clipping \cite{arjovsky2017towards}, gradient penalty (WGAN-GP) \cite{gulrajani2017improved}, batch norm \cite{ioffe2015batch}, layer norm \cite{ba2016layer}, weight norm \cite{salimans2016weight}, orthonormal regularization \cite{brock2016neural}, and spectral normalization (SN-GAN) \cite{spectral}. As can be seen, our models outperform the others by a large gap. Most notably, SAN-BigGAN performs substantially better than the original BigGAN, and sets a new state-of-the-art in conditional image generation on CIFAR-10.  

\subsection{Image-to-Image Translation}
Next, we illustrate our method in the challenging task of translating images between different domains. Here we focus on converting semantic segmentation masks to photo-realistic images. In the Appendix, we also demonstrate the power of SAN for attribute transfer.

We adopt the state-of-the-art SPADE scheme \cite{spade} as a baseline framework, and enhance its results by applying our normalization. We use the same multi-scale discriminator as \cite{spade}, except that we replace the zero padding by circular padding and preform SAN. To reduce the memory footprint, we use $r=25\%$ of the filters with a compensation factor of $g = 1.3$. All hyper-parameters are kept as in \cite{spade}, except for Adam's first momentum parameter, which we set to $\beta_1=0.5$.

\begin{figure}[t]
	\centering
	\includegraphics[width=0.9\columnwidth]{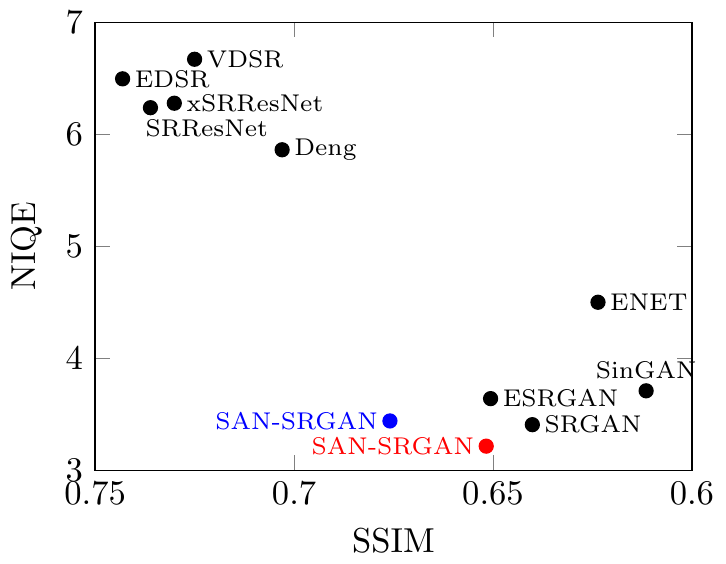}
	\caption{\textbf{Perception-distortion evaluation for SR.} We compare our models ($\lambda_{\text{adversarial}}= \color{red} 10^{-1}$ and $\color{blue} 10^{-2}$) to other state-of-the-art super-resolution models in terms of perceptual quality (NIQE, lower is better) and distortion (SSIM). Our method improves upon all existing  perceptual methods (those at the bottom right) in both perceptual quality and distortion.} 
	\label{fig:tradeoff}
\end{figure}

\addtolength{\tabcolsep}{3pt} 
\begin{table}[t]
	\begin{center}
		\begin{tabular}{p{1.8cm} @{\hskip3pt}c@{\hskip3pt} @{\hskip3pt}c@{\hskip3pt} @{\hskip3pt}c@{\hskip3pt}}
			Method & BSD100 & URBAN100 & DIV2K \\ \hline \hline
			
			SRGAN & 25.18 / 3.40 &  & \\
			ENET & 24.93 / 4.52 & 23.54 / 3.79 & \\
			ESRGAN & 25.31 / 3.64 & 24.36 / 4.23 & 28.18 / 3.14 \\
			ESRGAN* & 25.69 / 3.56 & 24.36 / 3.96 & 28.22 / 3.06 \\
			Ours ($\color{red} 0.1 $) & 25.32 / \textbf{3.21} & 23.86 / \textbf{3.70} & 27.74 / \textbf{2.87} \\
			Ours ($\color{blue} 0.01 $) & \textbf{26.15} / 3.44 & \textbf{24.85} / 3.83 & \textbf{28.76} / 3.16 \\
		\end{tabular}
	\end{center}
	\caption{PSNR/NIQE comparison among different perceptual SR methods on varied datasets. Our models attain a significantly higher PSNR and lower NIQE than other perceptual SR methods.}
	\label{tab:sr}
\end{table}
\addtolength{\tabcolsep}{-3pt} 

\begin{figure*}[h]
	\centering
	\includegraphics[width=2.0\columnwidth]{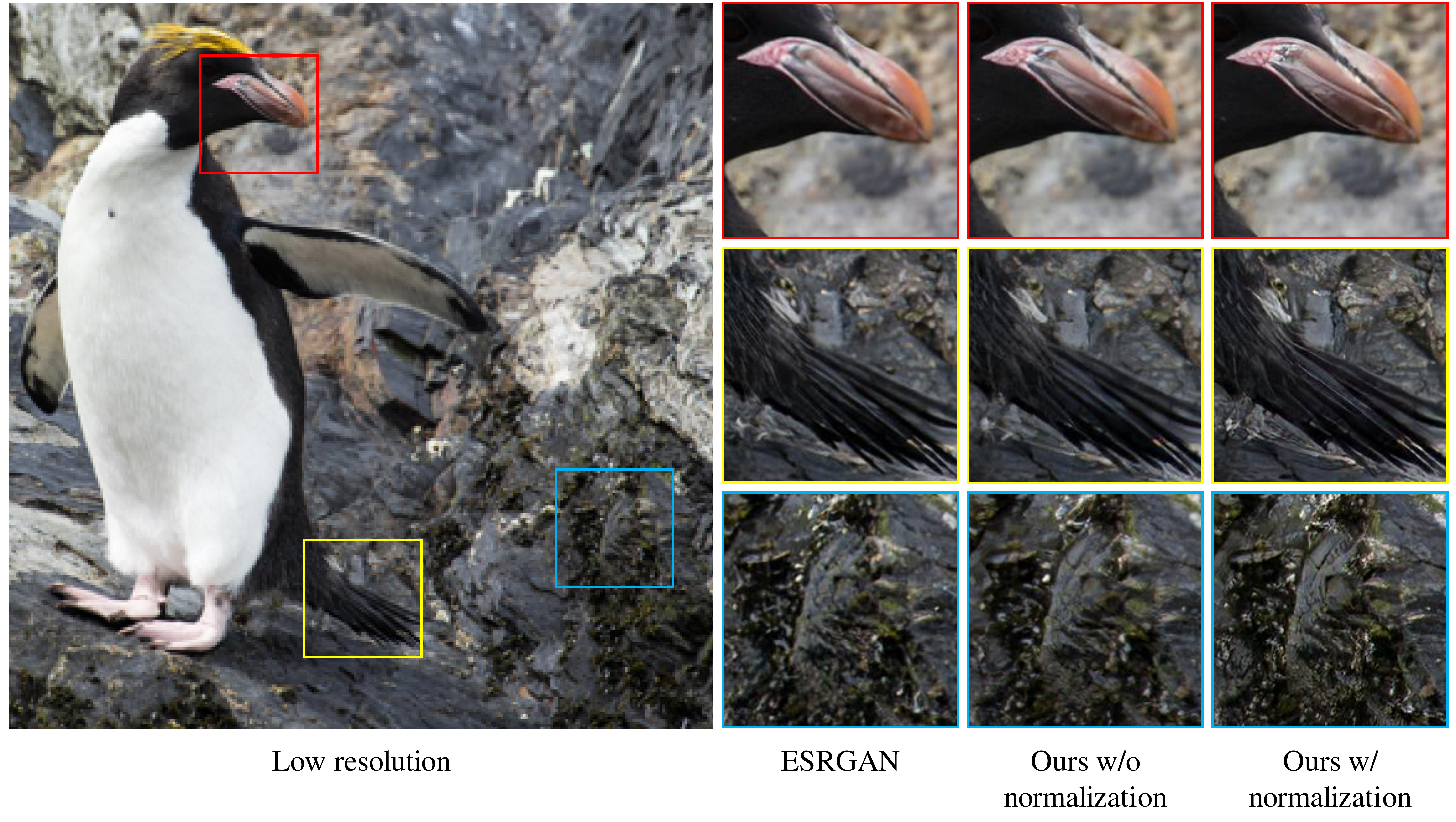}
	\caption{\textbf{The influence of normalization in super-resolution.} We compare the state-of-the-art ESRGAN method to our approach, with and without normalization, at a magnification factor of $4\times$. As can be seen, our normalization leads to sharper and more photo-realistic images.}
	\label{fig:sr_norm}
\end{figure*}

We use $512\times 256$ images from the Cityscapes dataset \cite{cityscapes}. For quantitative evaluation, we use the Fr\'echet Inception distance (FID). As can be seen in Fig.~\ref{fig:fids}, our method converges faster and leads to a final model that outperforms the original SPADE by a non-negligible margin. Specifically, SAN-SPADE achieves an FID of $58.56$ while the original SPADE achieves $63.65$. Figure~\ref{fig:image2image_spade} shows a qualitative comparison between SPADE and our SAN version after $1.1\times10^4$ iterations. As can be seen, our synthesized images have less artifacts and contain more details.

\subsection{Single Image Super Resolution}

Finally, we illustrate SAN in single image super resolution (SR), where the goal is to restore a high resolution image from its down-sampled low resolution version. We focus on  $4 \times$ SR for images down-sampled with a bicubic kernel.

Following the state-of-the-art ESRGAN \cite{esrgan} method, our loss function comprises three terms,
\begin{equation}\label{eq:sr_loss}
\mathcal{L} = \lambda_{\text{content}} \cdot \mathcal{L}_{\text{content}} + \mathcal{L}_{\text{features}} + \lambda_{\text{adversarial}} \cdot \mathcal{L}_{\text{adversarial}}.
\end{equation}
Here, $\mathcal{L}_{\text{content}}$ is the $L_1$ distance between the reconstructed high-res image~$\hat{x}$ and the ground truth image $x$. The term $\mathcal{L}_{\text{features}}$ measures the distance between the deep features of $\hat{x}$ and $x$, taken from $4^{\text{th}}$ convolution layer (before the $5^{\text{th}}$ max-pooling) of a pre-trained 19-layer VGG network \cite{vgg}. Lastly, $\mathcal{L}_{\text{adversarial}}$ is an adversarial loss that encourages the restored images to follow the statistics of natural images. Here, we use again the hinge loss.

\begin{figure*}[h]
	\centering
	\includegraphics[width=2.0\columnwidth]{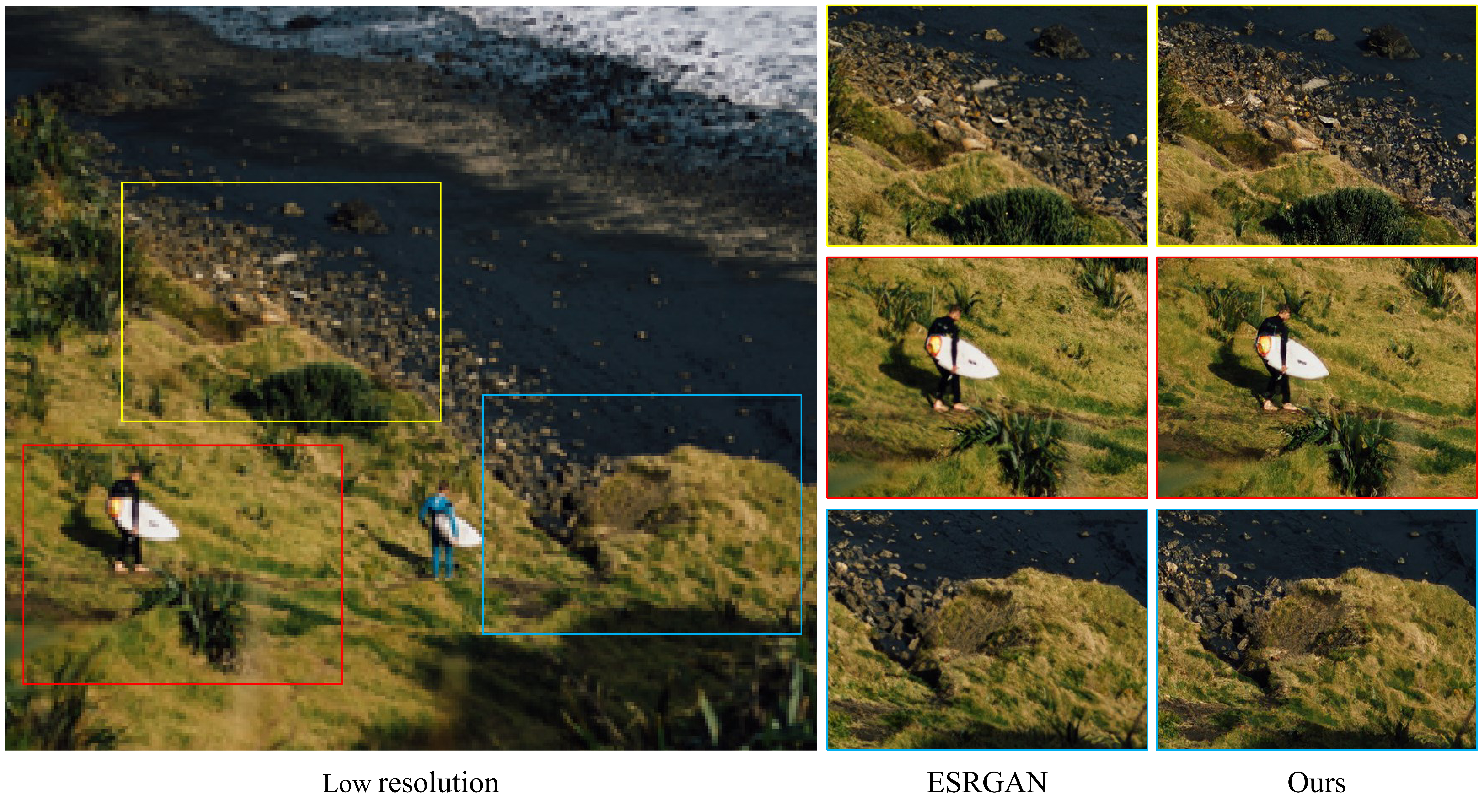}
	\caption{\textbf{Further super-resolution comparisons.} Compared to ESRGAN, our
method better recovers textures, like grass and stones.}
	\label{fig:sr2}
\end{figure*}

For the generator, we use the much slimmer SRGAN network \cite{srgan}, so that our model has only $9\%$ the number of parameters of ESRGAN ($1.5$M for ours and $16.7$M for ESRGAN). As suggested in \cite{edsr}, we remove the batch normalization layers from the generator. For the critic network, we choose a simple feed forward CNN architecture with 10 convolutional layers and 2 fully connected ones (ssee Appendix J).

We train our network using the 800 DIV2K training images \cite{div2k}, enriched by random cropping and horizontal flipping. The generator's weights are initialized to those of a pre-trained model optimized to minimize mean squared error. We minimize the loss \eqref{eq:sr_loss} with $\lambda_{\text{content}}=10^{-2}$, and for the adversarial term, $\lambda_{\text{adversarial}}$ we examine two options of $10^{-1}$ and $10^{-2}$. We use the Adam optimizer \cite{adam} with momentum parameters set to $0.5$ and $0.9$, as in Section \ref{image_generation}. We use a batch size of $32$ for $400$K equal discriminator and generator updates. The learning rate is initialized to $2\cdot 10^{-4}$ and is decreased by a factor of 2 at $12.5\%$,  $25\%$, $50\%$ and $75\%$ of the total number of iterations.

Following \cite{Blau_2018_CVPR}, we compare our method to other super-resolution schemes in terms of both perceptual quality and distortion. Figure \ref{fig:tradeoff} shows a comparison against EDSR \cite{edsr}, VDSR \cite{vdsr}, SRResNet \cite{srgan}, xSRResNet \cite{xunit}, Deng \cite{deng2018enhancing}, ESRGAN\cite{esrgan}, SRGAN \cite{srgan}, ENET \cite{enet} and SinGAN \cite{shaham2019singan}. Here, perceptual quality is quantified using the no-reference NIQE metric \cite{niqe} (lower is better), which has been found  in \cite{prim} to correlate well with human opinion scores in this task. Distortion is measured by SSIM \cite{ssim} (higher is better). We report average scores over the BSD100 test set \cite{bsd}. As can be seen, our method achieves the best perceptual quality, and lower distortion levels than the perceptual methods (at the bottom right).

In Table \ref{tab:sr}, we report comparisons to the best perceptual methods on two more datasets, the URBAN100 and DIV2K test sets. As the original ESRGAN \cite{esrgan} uses gradient penalty as a normalization scheme, for a fair comparison, we train an equivalent version, ESRGAN*, with spectral normalization \cite{spectral}. Note that our model outperforms ESRGAN (winner of the PRIM challenge on perceptual image super-resolution \cite{prim}) as well as the improved variant ESRGAN*. This is despite the fact that our generator network has only $9\%$ the number of parameters of ESRGAN's generator. Furthermore, while our model has the same generator architecture as SRGAN, it outperforms it by $1$dB in PSNR without any sacrifice in perceptual score.

Figures \ref{fig:sr} and \ref{fig:sr2} shows a visual comparison with ESRGAN. As can been seen, our method manages to restore more of the fine image details, and produces more realistic textures. Figure \ref{fig:sr_norm} shows yet another visual result, where we specifically illustrate the effect of our normalization. While without normalization our method is slightly inferior to ESRGAN, when we incorporate our normalization, the visual quality is significantly improved.

\subsection{Limitations}
SAN does not provide a boost in performance when the critic's feature maps do not exhibit strong channel-sparsity. This happens, for example, in BigGAN for $128\times128$ images (see Appendix H). There, there is one set of features in each res-block that are less sparse (those after the residual connection). A possible solution could be to use a different compensation factor $g$ for different layers, according to their level or sparsity. However, we leave this for future work.

\section{Conclusion}
We presented a new per-layer normalization method for GANs, which explicitly accounts for the statistics of signals that enter each layer. We showed that this approach stabilizes the training and leads to improved results over other GAN schemes. Our normalization adds a marginal computational burden compared to the forward and backward passes, and can even be applied once every several hundred steps while still providing a significant benefit.\\

\noindent\textbf{Acknowledgements\ \ \ }
This research was supported by the Israel Science Foundation (grant 852/17) and by the Technion Ollendorff Minerva Center.

\bibliography{egbib}
\newpage
\appendixpage
\appendix

\section{Proof of Example~1}
Consider the function $f:\mathbb{R}\rightarrow\mathbb{R}$ defined by
\begin{equation}
	f(x)=w_2^T \sigma(w_1 x + b_1) + b_2,
\end{equation}
where $\sigma$ is the ReLU activation function, $w_1,w_2,b_1\in\mathbb{R}^n$, and $b_2\in\mathbb{R}$. Our goal is to characterize the slopes of this function at $+\infty$ and $-\infty$, assuming that $\|w_1\|\leq 1$ and $\|w_2\|\leq 1$.

Writing the function explicitly in terms of the entries of the parameter vectors, we obtain
\begin{equation}
f(x)=\sum_{i=1}^n (w_2)_i \cdot\max\left\{(w_1)_i\, x + (b_1)_i,0\right\} + b_2.
\end{equation}
Therefore, 
\begin{equation}
f'(x)=\sum_{i\in\mathcal{I}_x}(w_1)_i (w_2)_i,
\end{equation}
where
\begin{equation}
\mathcal{I}_x=\{i:(w_1)_i\,x+(b_1)_i>0\}.
\end{equation}
Now, note that 
\begin{align}
\mathcal{I}_\infty=\{i:(w_1)_i>0\}, \qquad
\mathcal{I}_{-\infty}=\{i:(w_1)_i<0\}.
\end{align}
We thus have that
\begin{align}
|f'(-\infty) + f'(\infty)| &= \left|\sum_{i\in\mathcal{I}_{-\infty}}(w_1)_i (w_2)_i + \sum_{i\in\mathcal{I}_{\infty}}(w_1)_i (w_2)_i\right| \nonumber\\
&= \left|\sum_{i=1}^n(w_1)_i (w_2)_i\right| \nonumber\\
&\leq  \|w_1\|\, \|w_2\| \nonumber\\
&\leq 1,
\end{align}
where in the second line we added to the sum all indices $i$ for which $(w_1)_i=0$ (the sum over these indices gives zero), in the third line we used Parseval's inequality, and in the last line we used the fact that $\|w_1\|\leq 1$ and $\|w_2\|\leq 1$.

\section{Proof of Lemma~1}
For a single-input-single-output cyclic convolution operation, $y=h*x$, it is well known that the top singular value is $\|\mathcal{F}\{h\}\|_\infty$, the maximal absolute value of the discrete Fourier transform of $h$ (after zero-padding it to the length of $x$). Now, for a multiple-input-multiple-output convlution $W$, we have
\begin{align}
\|W\|_{0,\infty} &= \sup_{\|\boldsymbol{x}\|_0 \leq 1\atop\|\boldsymbol{x}\|_\infty \leq 1} \|W\boldsymbol{x}\|_\infty \nonumber\\
&= \sup_{\|\boldsymbol{x}\|_0 \leq 1\atop\|\boldsymbol{x}\|_\infty \leq 1} \max_{i}\Big\|\sum_j w_{i,j}*x_j\Big\| \nonumber\\
& =  \sup_{\|\boldsymbol{x}\|_\infty \leq 1} \max_j \max_{i}\| w_{i,j}*x_j\| \nonumber\\
& =  \max_j \max_{i} \sup_{\|x_j\| \leq 1}\| w_{i,j}*x_j\| \nonumber\\
& =  \max_j \max_{i} \|\mathcal{F}\{w_{i,j}\}\|_\infty,
\end{align}
where in the third line we used the fact that when $\|\boldsymbol{x}\|_0 \leq 1$, the sum contains only one nonzero element, and in the last line we used the top singular value of a single-input-single-output convolution.

\begin{figure*}[h]
\begin{center}
\begin{tabular}{ p{8cm} c }  
\vspace{-1.75cm}\includegraphics[width=0.4\textwidth]{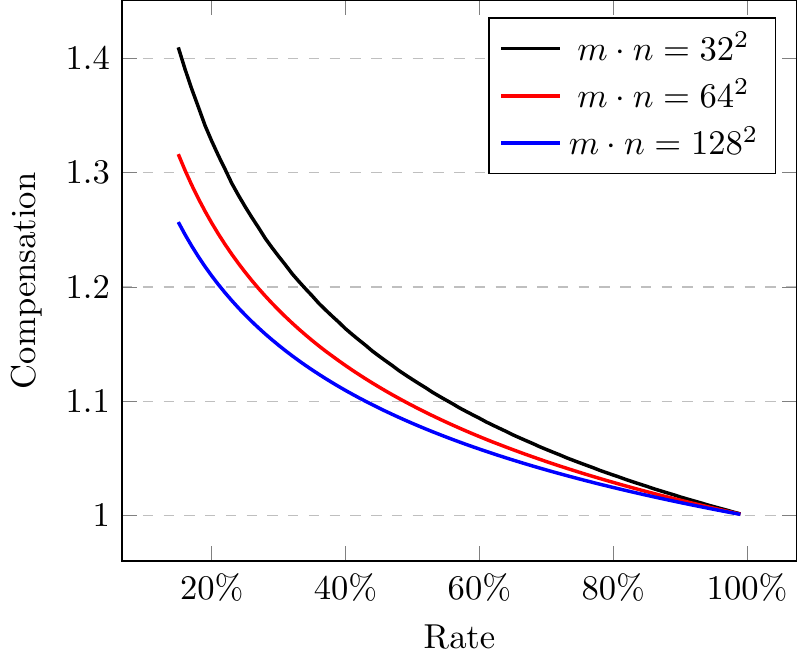}
&
\renewcommand{\arraystretch}{2}
\begin{tabular}{ll|l|l|l|l|}
        
\multicolumn{2}{c}{}&   \multicolumn{4}{c}{Compensation}\\
\multicolumn{2}{c}{}&\multicolumn{4}{c}{
              {\rotatebox[origin=c]{0}{    1.0    }
            } {\rotatebox[origin=c]{0}{    1.1    }
            } {\rotatebox[origin=c]{0}{    1.2    }
            } {\rotatebox[origin=c]{0}{    1.3    }
        }}\\
\cline{3-6}
\multirow{3}{*}{{\rotatebox[origin=c]{90}{Rate}
}} & 
100\% & 73.8 & \textbf{65.8} & 78.8 & 69.6  \\ \cline{3-6}
&   50\%  & 76.6 & 73.6 & \textbf{72.4} & 79.3 \\ \cline{3-6}
&   25\%  & 74.3 & 74.7 & 75.2 & \textbf{67.8} \\ \cline{3-6}
\end{tabular}\\
& \\
\end{tabular}
\caption{\textbf{Compensation factor.} The left pane depicts the theoretical compensation factor $g$ for varied filter sizes, as a function of the percentage of filters $r$ used to compute the normalization constant. The table on the right details the Fr\'echet Inception distance (FID) attained in Cityscapes label-to-image translation, for different ratios and compensation factors. As can be seen, the optimal compensation factor indeed increases as the ratio of filters decreases.}
\label{fig:study_compensation}
\end{center}
\end{figure*} 

\begin{algorithm*}
\caption{SGD with sparsity aware normalization (SAN))}\label{alg:san}
\begin{algorithmic}[1]
\For {each layer, $\ell = 1, \dots, n$}

\State Calculate $\sigma$ from a random subset of filters $\{\tilde{w}^\ell_{i,j}\} \subset \{w^\ell_{i,j}\}$,
\begin{align*}
\sigma = \max_{i,j}\|\mathcal{F}\{\tilde{w}_{i,j}\}\|_\infty.
\end{align*}
\State Normalize $W^\ell$ using a compensation factor $g$,
\begin{align*}
W^\ell \leftarrow W^\ell / (g \cdot \sigma) .
\end{align*}
\State Update $W^\ell$ using a mini-batch of $M$ samples with a learning rate of $\alpha$, as
\begin{equation*}
W^\ell \leftarrow W^\ell - \alpha \tfrac{1}{M}\sum_{i=1}^{M}\nabla_{W^\ell}\mathcal{L}(x_i,y_i), 
\end{equation*}
where $\mathcal{L}(x_i,y_i)$ is the critic loss for input $x_i$ and label $y_i$.

\EndFor

\end{algorithmic}
\end{algorithm*}

\section{The Compensation Factor}
\begin{figure}[h!]
\centering
\includegraphics[width=0.8\columnwidth]{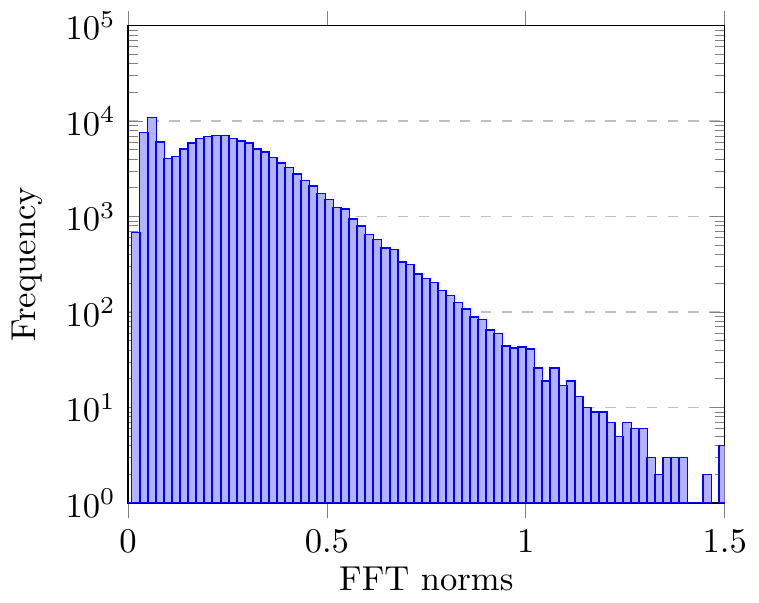}
\caption{\textbf{Exponential decay}. The histogram of values $\{ \|\mathcal{F}\{w_{i,j}\}\|_\infty\}$ for the last layer of the standard critic network used in the paper.}
\label{fig:hist_ffts}
\end{figure}

Next, we study the compensation factor $g$ that has to be used when computing our normalization constant based on only a (random) subset of the filters. Figure~\ref{fig:hist_ffts} depicts the histogram of values $\{ \|\mathcal{F}\{w_{i,j}\}\|_\infty\}$ for the last layer of the standard critic network used in the paper. This layer has  $m=256$ input channels and $n=512$ output channels, so that the histogram is over  $256\times 512\approx 130,000$ values. As can be seen, the decay of the histogram is roughly exponential. This suggests that the values $\{ \|\mathcal{F}\{w_{i,j}\}\|_\infty\}$ can be regarded as realizations of independent random variables $\{X_j\}_{j=1}^{m\cdot n}$ distributed $\text{exp}(\lambda)$, for some $\lambda>0$. Now, assuming we compute the maximum over only $k<m\cdot n$ values, we would like to know how much smaller the result would be on average. Namely, we would like to compare $\E[\max \{X_j\}_{j=1}^k]$ with $\E[\max \{X_j\}_{j=1}^{m\cdot n}]$.

In \cite{renyi1953theory}, Alfr\'ed R\'enyi showed that the $i$th order statistics of $k$ iid samples from an exponential distribution is given by
\begin{equation}
X_{(i)} \overset{d}{=}  \frac{1}{\lambda} \left(\sum_{j=1}^{i} \frac{Z_j}{k-j+1}\right),
\label{eq:order_distribution}
\end{equation}
where $\overset{d}{=}$ denotes equality in distribution and the $Z_j$ are iid standard exponential random variables (with $\lambda=1$). Thus, focusing on the maximum, which corresponds to $i=k$, and taking the expectation of the expression, we obtain
\begin{equation}
\E \left[\max\{X_j\}_{j=1}^k\right] = \frac{1}{\lambda} \left(\sum_{j=1}^{n} \frac{1}{n-j+1}\right),
\label{eq:order_maxium}
\end{equation}
Therefore, when taking a ratio of $r$ of the filters (so that $k=\round{m \cdot n \cdot r}$, the compensation factor should be taken to be
\begin{equation}
g=\frac{\E \left[\max\{X_j\}_{j=1}^{m\cdot n}\right]}{\E \left[\max\{X_j\}_{j=1}^{\round{m \cdot n \cdot r}}\right]}=\frac{\sum_{j=1}^{m\cdot n} \frac{1}{m\cdot n-j+1}}{\sum_{j=1}^{\round{m \cdot n \cdot r}} \frac{1}{\round{m \cdot n \cdot r}-j+1}}.
\label{eq:compensation_factor}
\end{equation}

Interestingly, this value varies quite slowly with $r$ for large $m$ and $n$. This can be seen in Fig.~\ref{fig:study_compensation} (left), where we depict the theoretical optimal value $g$ of \eqref{eq:compensation_factor} for varied filters sizes. For example, $g$ typically does not exceed $1.3$ even for ratios as low as $25\%$.

To verify Eq.~\eqref{eq:compensation_factor}, we also conduct an experiment on the task of Cityscapes label-to-image translation task. For this objective, we adopt the SPADE \cite{spade} framework with narrower generator and discriminator networks as a baseline, apply SAN in the discriminator and train several models with different ratios $r$ and compensation factors $g$. In Fig.~\ref{fig:study_compensation} (right), we show the Fr\'echet Inception distances (FID) attained for each model. The best result is attained with a rate and compensation of $100\%$ and $1.1$, respectively. Furthermore, the practical optimal compensation factor roughly agrees with the theoretical one for rates lower than $100\%$.

\section{The Sparsity Aware Normalization Algorithm}

Algorithm \ref{alg:san} describes the SAN algorithm update of the critic within an SGD step. In models for large images, we compute the normalization constant over a random subset of the filters as is determined by Eq.~\eqref{eq:compensation_factor}.

\section{Ablation Study}

\begin{figure}[t]
\centering
\includegraphics[width=0.75\columnwidth]{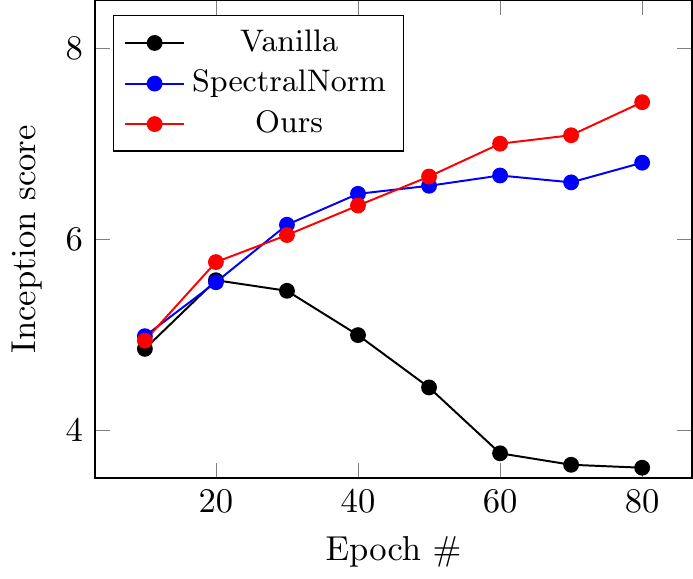}
\caption{\textbf{Convergence}. Inception score on CIFAR-10 during training with three WGANs different methods.}
\label{fig:convergence}
\end{figure}

In this section, we explore and evaluate our sparsity aware normalization on a series of experiments. We focus on image generation for CIFAR-10 with the standard CNN architecture, and compare the inception score (IS) of our SAN-GAN to those of spectral normalization (SN-GAN) \cite{spectral} and an un-normalized GAN variant (Vanilla). In all the experiments, the number of updates for the generator is set to $80$K, and all methods are initialized with the same draw of random weights

To explore how the usage of our normalization impacts the optimization of WGANs, we measure the inception score (IS) of the three methods during training. As shown in Fig.~\ref{fig:convergence}, our method exhibits a faster and steadier convergence.

\begin{figure}[t]
\centering
\includegraphics[width=0.775\columnwidth]{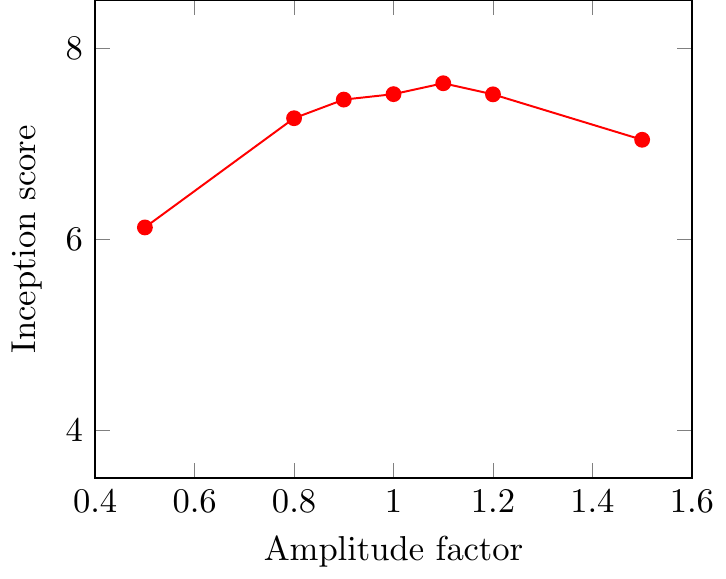}
\caption{\textbf{Looseness of normalization}. Inception score on CIFAR-10 for different $\ell_\infty$ norm constraints.}
\label{fig:looseness_norm}
\end{figure}

We continue by preforming an ablation study on the looseness of the normalization. As loosening the bound in our $\|\boldsymbol{\cdot}\|_0$ constraint is computationally impractical, we focus only on loosening the $\|\boldsymbol{\cdot}\|_\infty$ bound, which boils down to multiplying the normalization constant by some correction. Therefore, in practice, this implies using a different compensation factor $g$. In Fig.~\ref{fig:looseness_norm}, we vary the magnitude of the norm constraint between $0.5$ and $1.5$. Interestingly, we obtained the best results with a value of $1.1$, a small improvement over the results with $1.0$, which is proposed in the paper. This suggests that our approach, with a bound of $1.0$ does actually predict a rather good normalization
constant.

\begin{figure*}[h]
	\begin{center}
		\vspace{-3cm}
		\begin{tabular}{ p{7cm} c }  
			\vspace{0cm}\includegraphics[width=0.33\textwidth]{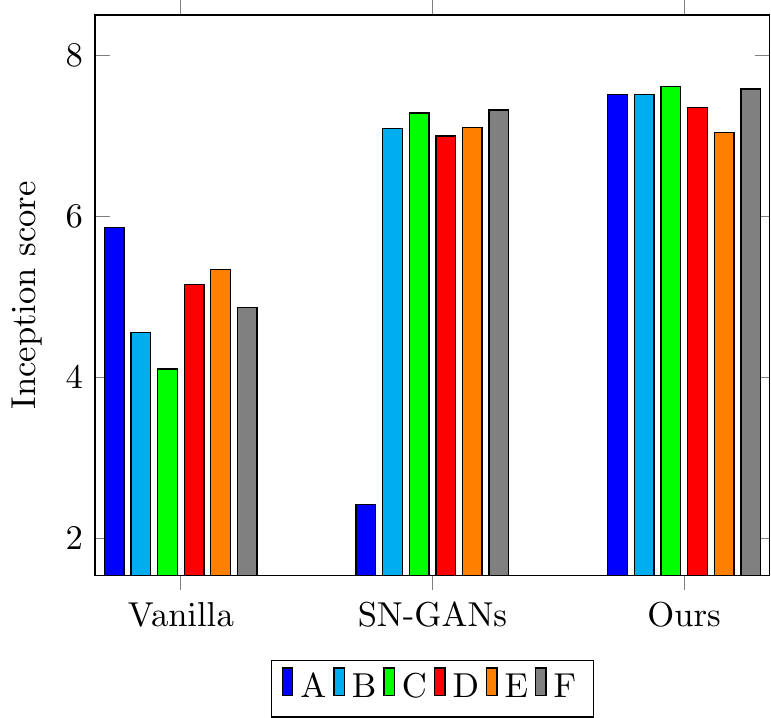}
			&
			\begin{tabular}{ L{1.2cm} C{1cm} C{0.6cm} C{0.6cm} C{0.6cm} }
				\\ \\ \\ \\ \\ \\ \\ \\ \\
				Setting & $\alpha$ & $\beta_1$ & $\beta_2$ & $n_{\text{dis}}$ \\ \hline
				A & $0.0005$ & $0.5$ & $0.9$ & $1$ \\
				B & $0.0002$ & $0.5$ & $0.9$ & $1$ \\
				C & $0.0002$ & $0$ & $0.9$ & $1$ \\
				D & $0.0005$ & $0$ & $0.9$ & $1$ \\
				E & $0.0002$ & $0.5$ & $0.9$ & $2$ \\
				F & $0.0005$ & $0$ & $0.9$ & $2$ \\
			\end{tabular}\\
			& \\
		\end{tabular}
		\caption{\textbf{Resistance to training hyper-parameters.} Inception scores on CIFAR-10 with different methods and training hyper-parameters.}
		\label{tab:study_confs}
	\end{center}
\end{figure*}

To characterize the dependence of our sparsity aware normalization on the training configuration, we inspect all three methods under six hyper-parameter settings (A-F) corresponding to different learning rates $\alpha$,  different first and second order momentum parameters $(\beta_1, \beta_2)$ for the Adam optimizer, and different numbers of critic updates for each generator update, $n_\text{dis}$. As can be seen in Fig. \ref{tab:study_confs}, our method outperforms spectral normalization in all settings except for E. This is true even with a high learning rate, where spectral normalization fails (set A). These results suggest that our method is more robust than the competing methods to perturbations in the training settings.

\begin{figure*}[h]
	\begin{center}
		\vspace{-3cm}
		\begin{tabular}{ p{10cm} c }  
			\vspace{0cm}\includegraphics[width=0.33\textwidth]{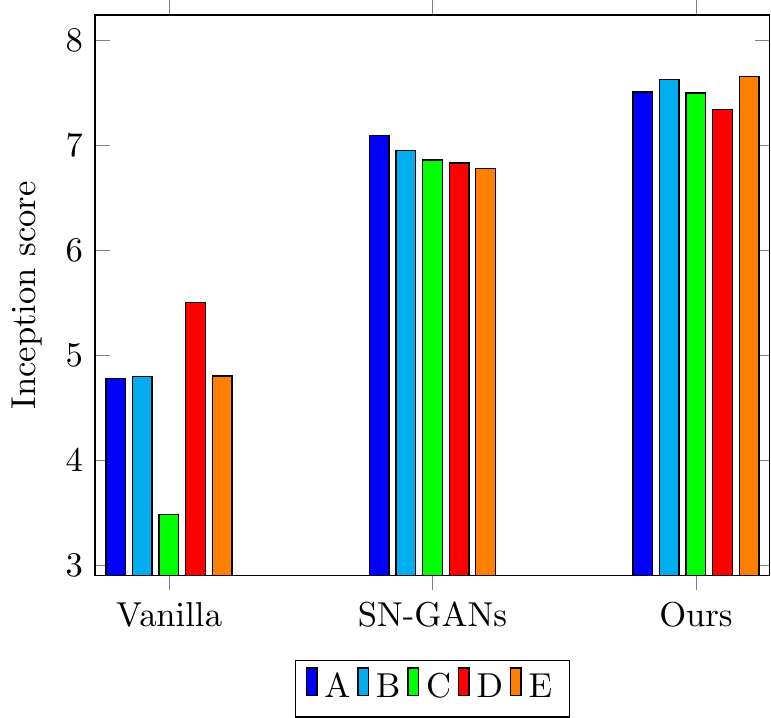}
			&
			\begin{tabular}{ L{1.2cm} C{1.2cm} }
				\\ \\ \\ \\ \\ \\ \\ \\ \\
				Setting & \# seed \\ \hline
				A & $5354$ \\
				B & $9017$ \\
				C & $11224$ \\
				D & $5577$ \\
				E & $61$ \\
			\end{tabular}\\
			& \\
		\end{tabular}
		\caption{\textbf{Robustness to weight initialization.} Inception scores on CIFAR-10 with different methods and initial weight sets.}
		\label{tab:study_seed}
	\end{center}
\end{figure*}

To see how sensitive the results are to the particular weight initialization, in Fig.~\ref{tab:study_seed} we show the inception scores of each method with set B for several different seeds of the random number generator. As can be seen, our method consistently outperforms the competitors, indicating that it is more robust to the specific draw of the weights at initialization.

\section{Running Time}
On CIFAR-10, SAN-GANs is slightly slower than running with no normalization ($\sim +15\%$ in computation time), is similar to spectral normalization ($\sim +2\%$ in computation time), but significantly faster than WGAN-GP ($\sim -17\%$ in computation time). Specifically, on our system, for one critic iteration update for the ``standard'' network (Table~\ref{tab:arch_gen_standard}), the average computation times are:
\begin{itemize}

\vspace{-0.05cm}

\item Vanilla: $35.74ms$
\vspace{-0.05cm}

\item WGAN-GP: $50.02ms$
\vspace{-0.05cm}

\item Spectral-Norm: $40.51ms$
\vspace{-0.05cm}

\item Ours: $41.45ms$

\end{itemize}

It is important to note, however, the $15\%$ overhead for one update typically does not translate to a $15\%$ overhead in the total training time. This is because, as discussed above, we do not have to use our normalization every update step and we can compute the normalization constant over only a random subset of filters. 

\section{Image-to-Image Translation}

\begin{table*}[t]
\begin{center}
\begin{tabular}{c c c c c c c }
\centering
Black & Blond & Brown & Female & Male & StarGAN & SAN-StarGAN \\ \hline
\ding{51} & \ding{55} & \ding{55} & \ding{51} & \ding{55} & 85.10 & \textbf{71.41}\\
\ding{51} & \ding{55} & \ding{55} & \ding{55} & \ding{51} & 88.72 &  \textbf{81.01}\\
			
\ding{55} & \ding{51} & \ding{55} & \ding{51} & \ding{55} & 189.65 & \textbf{184.82}\\
\ding{55} & \ding{51} & \ding{55} & \ding{55} & \ding{51} & 117.06 & \textbf{100.51}\\
			
\ding{55} & \ding{55} & \ding{51} & \ding{51} & \ding{55} & 101.33 & \textbf{87.91}\\
\ding{55} & \ding{55} & \ding{51} & \ding{55} & \ding{51} & 87.15 &  \textbf{76.32}\\
			
\end{tabular}
\end{center}
\caption{Quantitative comparison on the task of gender translation. The left 5 columns describe the source domain. The target domain corresponds to the same hair color, but the opposite gender. The Fr\'echet inception distance (FID) measures the discrepancy between the distributions of real and generated images (lower is better).}
\label{tab:image2image}
\end{table*}
\addtolength{\tabcolsep}{-3pt} 

Here, we provide additional image-to-image translation results for our method. Specifically, we use the CelebA dataset \cite{celeba} and focus on the task of modifying a particular attribute of an input image, like gender or hair color, to have some desired value, like male/female or black/blond/brown. We adopt StarGAN scheme \cite{stargan} as a baseline framework, and enhance its results by incorporating our normalization. As in \cite{stargan}, we take the generator to comprise two convolution layers with stride 2 (for downsampling), followed by six residual blocks, and two transposed convolution layers with stride 2 (for upsampling). For the critic, we use six convolution layers with circular padding, but reduce the kernel size from $4 \times 4$ to $3 \times 3$, so that the number of parameters in our model is only $25$M (in contrast to the $44$M of \cite{stargan}). Please see Tab. \ref{tab:arch_resnet_stargan} for more details about the network architectures. 

Like StarGAN, we optimize an objective function comprising three terms,
\begin{equation}\label{eq:im2im_loss}
\mathcal{L} = \lambda_{\text{cls}} \cdot \mathcal{L}_{\text{cls}} + \lambda_{\text{cycle}} \cdot \mathcal{L}_{\text{cycle}} + \mathcal{L}_{\text{adversarial}}.
\end{equation}
Here, $\mathcal{L}_{\text{cls}}$ is a domain classification loss, which ensures that the discriminator can properly classify the attributes of the generated images. The term $\mathcal{L}_{\text{cycle}}$ is a cycle consistency loss \cite{zhu2017unpaired}, and $\mathcal{L}_{\text{adversarial}}$ is a WGAN adversarial loss. But as opposed to StarGAN, which uses the plain Wasserstein loss together with a gradient penalty \cite{gulrajani2017improved}, here we use the hinge loss together with our sparsity-aware normalization. The hyper-parameters $\lambda_{\text{cls}}$ and $\lambda_{\text{cycle}}$ control the relative balance between the three terms, and as in \cite{stargan}, we set them to $\lambda_{\text{cls}}=1$ and $\lambda_{\text{cycle}}=10$. We minimize the loss \eqref{eq:im2im_loss} using the Adam optimizer \cite{adam} with $\beta_1=0.5$ and $\beta_2=0.9$. The batch size is set to $16$, as in \cite{stargan}. We train the networks with a learning rate of $10^{-4}$ for the first $10$ epochs and then apply a linear decay over the next $5$ epochs.

\begin{figure*}[thp]
\centering
\includegraphics[width=0.825\textwidth]{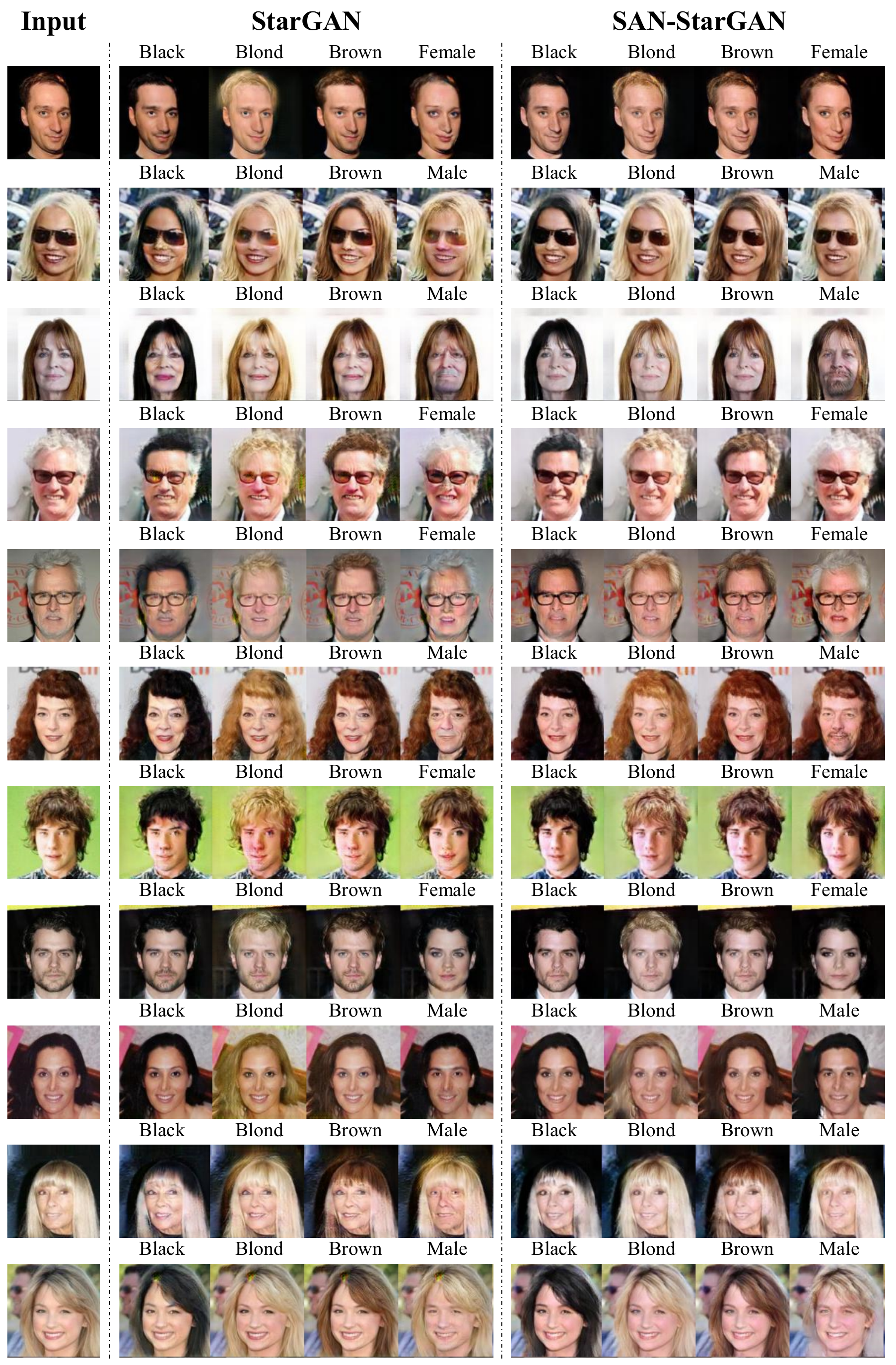}
\caption{\textbf{Visual comparison for image-to-image translation.} Each column corresponds to modifying a single attribute of the input image on the left. Our model manages to avoid artifacts (hallows around the head, colors on the face) and generate more photo-realistic images. Moreover, as opposed to the original StarGAN, our model does not modify attributes other than the desired one. For example, it does not add facial hair (as in row 1, column 1), fade sunglasses (as in row 2, columns 2,4), or change the apparent age (as in row 3, columns 1,2,3) when unnecessary.}
\label{fig:image2image_star}
\end{figure*}

Figure~\ref{fig:image2image_star} shows a qualitative comparison between StarGAN and our sparsity-aware normalized version, SAN-StarGAN. As can be seen, our images contain less artifacts, such as hallows around the hair or unrealistic colors on the face. Furthermore, our model better maintains attributes that do not need to change. For example, as opposed to StarGAN, it does not add a trace of a beard in row 2, and it does not cause the sunglasses to fade in row 4.

For quantitative evaluation, we use the 2K test images to generate 10K fake images by translating each image to one of five domains: `black hair', `blond hair', `brown hair', `gender' and `aged'. We compute the Fr\'echet Inception distance (FID) \cite{heusel2017gans} between the translated images and the training images in the target domain. In Table \ref{tab:image2image}, we summarize the scores for the challenging task of gender translation. As can been seen, the translated images produced by our model have greater statistical similarity to the desired target domain.

\section{Large Scale Image Generation}

As mentioned in the paper, SAN does not provide a boost in performance when the critic's feature maps do not exhibit a strong channel-sparsity, which is a crucial assumption underlying our theoretical motivation. For large scale image generation, ResNet has become the architecture of choice. However, compared to feed-forward architectures, the residual connections in ResNet suppresses sparsity in certain points along the path.

\begin{figure}[h!]
\centering
\includegraphics[width=0.6\columnwidth]{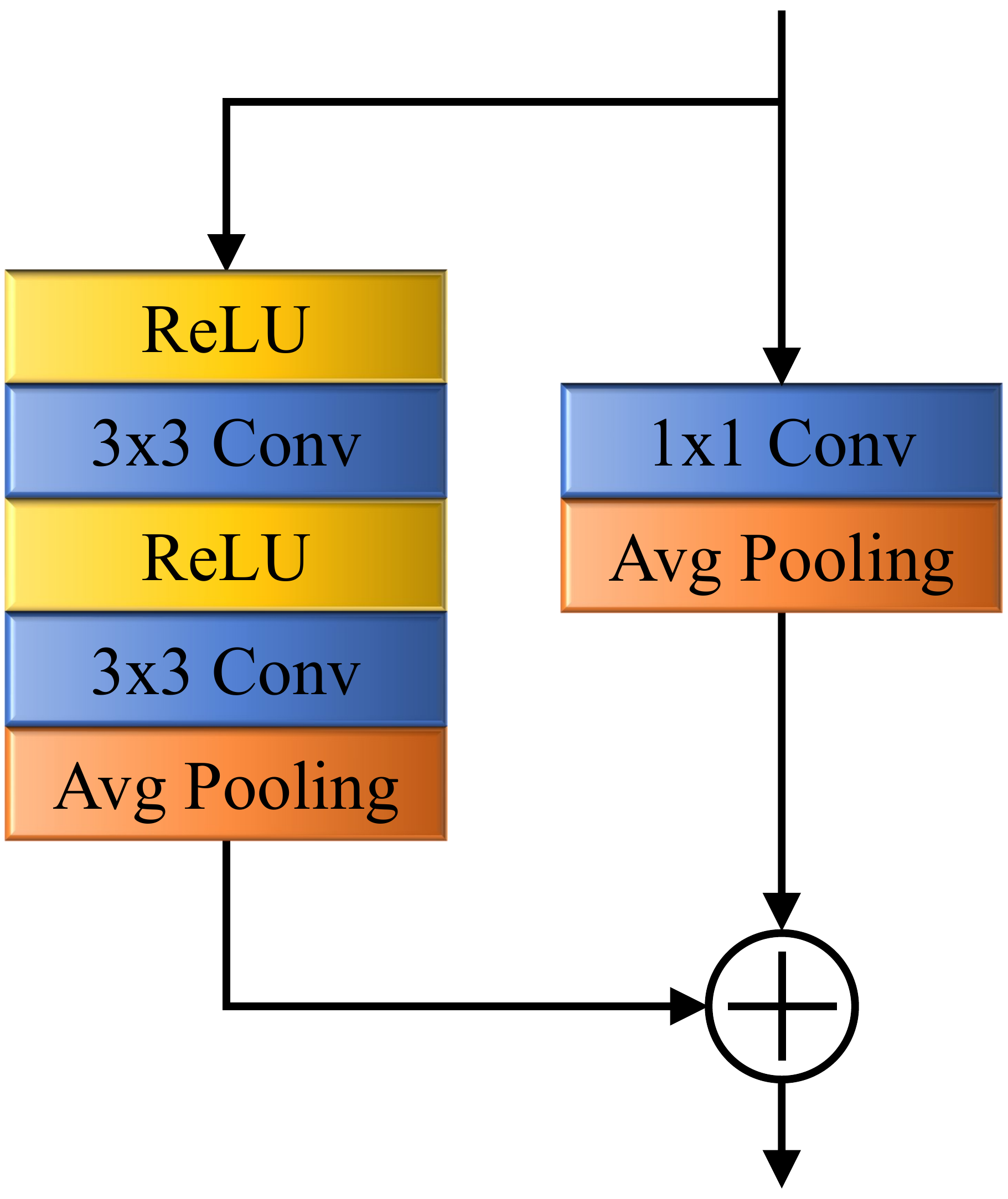}
\caption{\textbf{A BigGAN's discriminator ResBlock}.}
\label{fig:BigGAN_ResBlock}
\end{figure}

\begin{figure*}[thp]
\begin{center}
\begin{tabular}{ p{2.75cm} c c}
\vspace{-3.5cm}
$3rd$ Residual Block &
\includegraphics[scale=0.75]{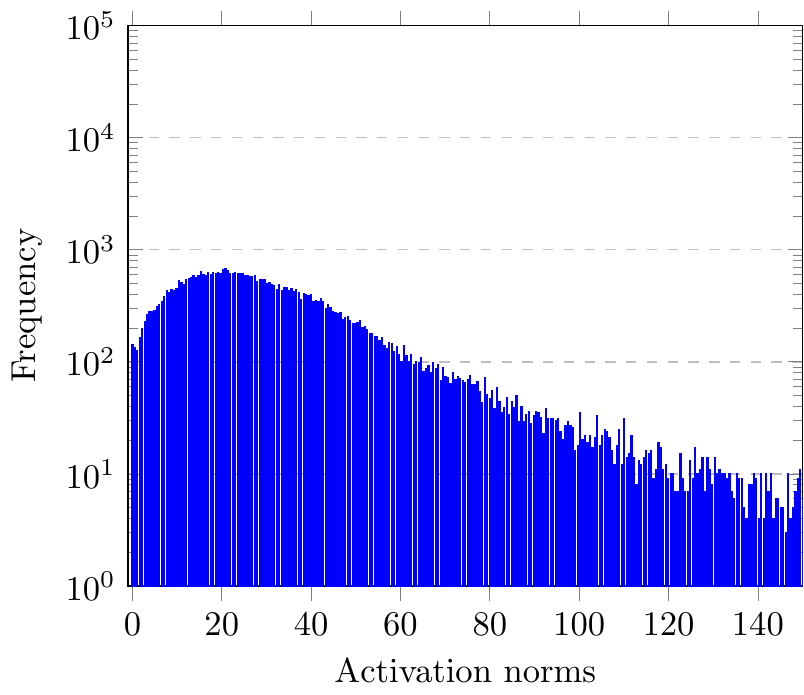} &
\includegraphics[scale=0.75]{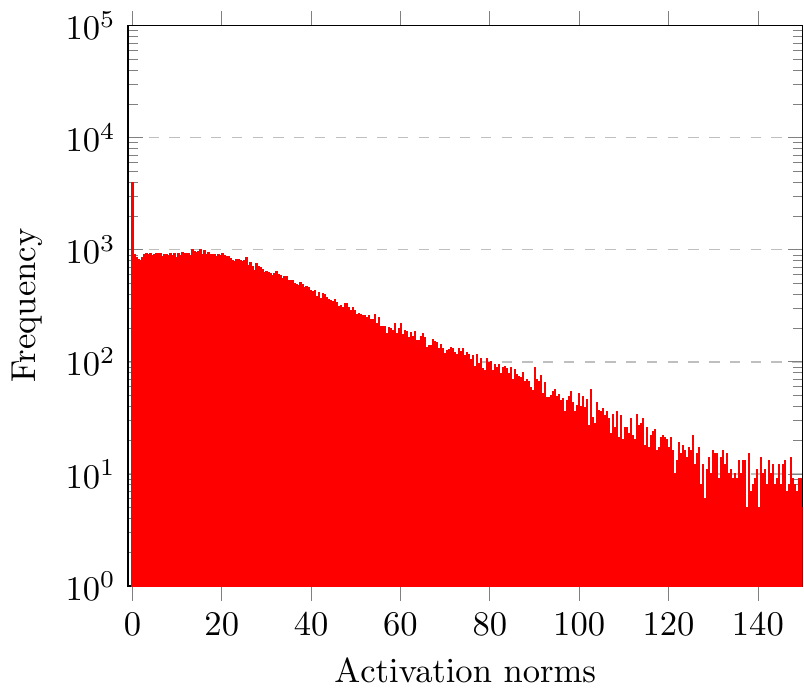} \\

\vspace{-3.5cm}
$4th$ Residual Block &
\includegraphics[scale=0.75]{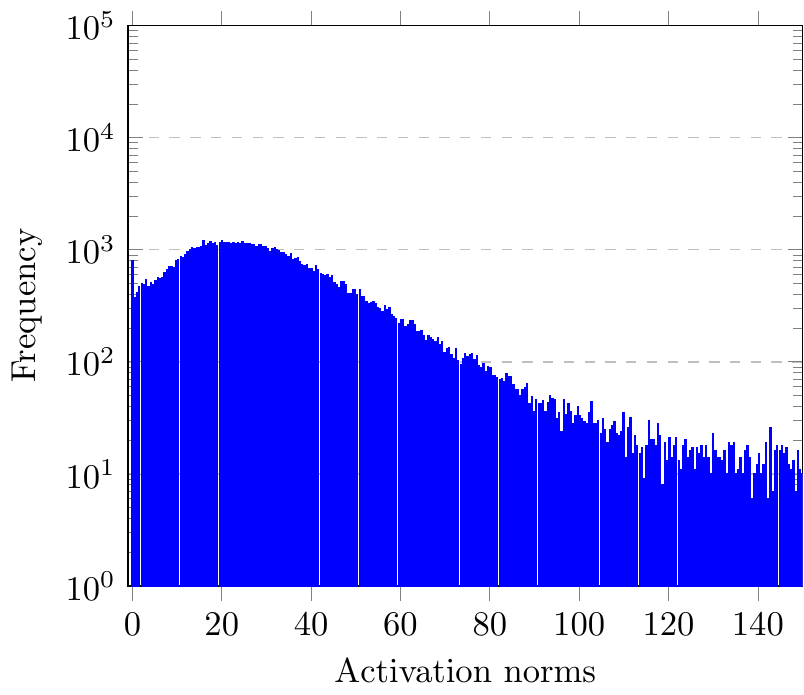} &
\includegraphics[scale=0.75]{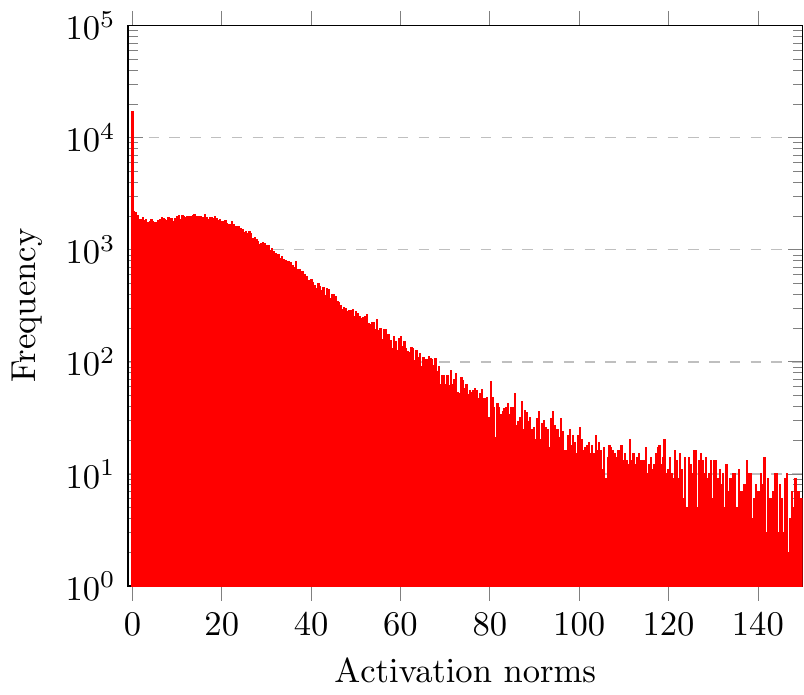} \\

\vspace{-3.5cm}
$5th$ Residual Block &
\includegraphics[scale=0.75]{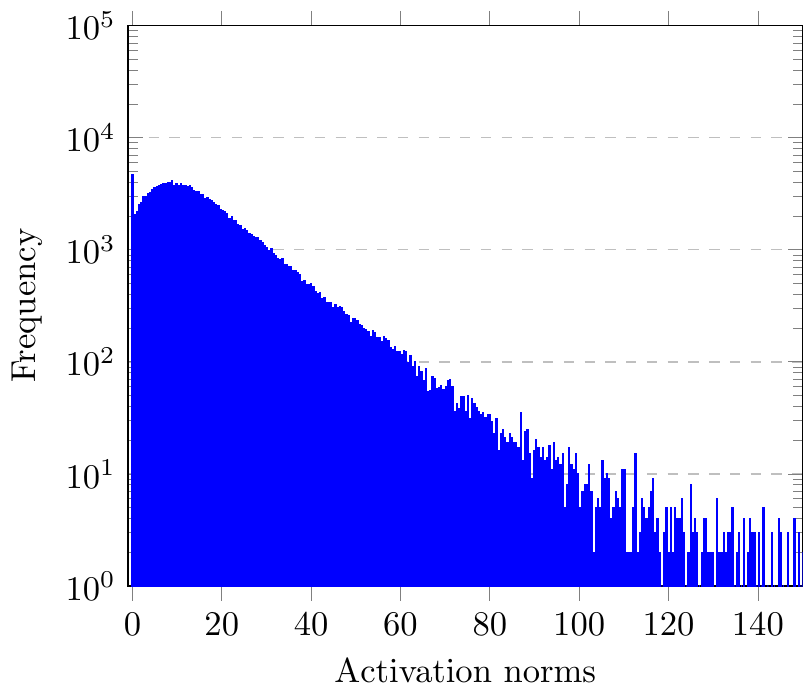} &
\includegraphics[scale=0.75]{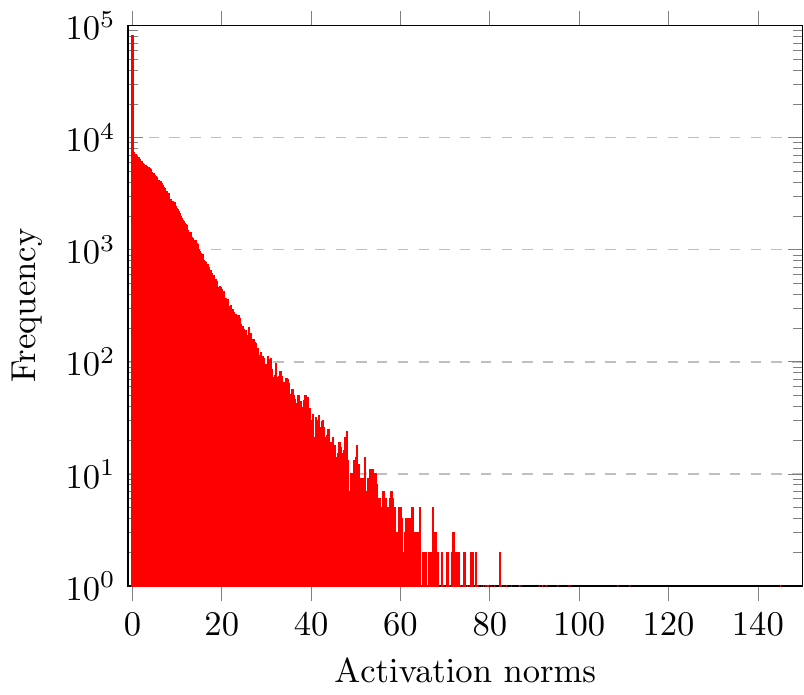} \\ \\
&
\hspace{1cm}
1st Convolution &
\hspace{1cm}
2nd Convolution

\end{tabular}
\caption{\textbf{Channel sparsity in BigGAN.} The histograms of the channel norms for several convolutional layers in a pre-trained BigGAN critic network, computed over $256$ randomly chosen samples from the ImageNet training set. The inputs to the first convolution (blue) are less sparse than the inputs to the second convolution (red). Note particularly the sharp peak at 0 in the red histograms.}
\label{tab:sparse_biggan}
\end{center}
\end{figure*}

To support this claim, in Fig.~\ref{tab:sparse_biggan}, we plot the histogram of norms of features for the pre-trained BigGAN critic \cite{brock2018large}, computed over $256$ randomly sampled $128\times 128$ images from the validation set within several residual blocks. In BigGAN, each ResBlock has two spatial convlutional layers (see Fig.~\ref{fig:BigGAN_ResBlock}). Interestingly, the inputs of the first convolutional layer in each ResBlock are not sparse, whereas the inputs of the second convolutional layer have some sparsity characteristics. This implies that the residual connection here eliminates sparsity. A possible solution could be to move the first ReLU so it would include the residual connection as commonly done in ResNets for image classification \cite{resnets}. Another possible solution may be to use a different compensation factor $g$ for different layers, according to their level of sparsity. We leave these for future work. Thus currently, our method is more applicable in feed-forward critic architectures, which actually dominate in image-to-image translation tasks.

\section{Additional Super-Resolution Results}
In Fig.~\ref{fig:sr} we show additional super-resolution results achieved with our approach, and compare them to the state-of-the-art ESRGAN.

\section{Experimental Setting}
In Tables \ref{tab:arch_gen_standard},\ref{tab:arch_resnet_cifar},\ref{tab:arch_resnet_stl},\ref{tab:arch_resnet_stargan} and \ref{tab:arch_sr}, we provide the exact network configuration for each of the experiments in the paper. In all implementations, we use cyclic padding for the convolutional layers of the critic. 

\newpage

\newpage

\newpage

\begin{figure*}[thp]
	\centering
	\vspace{-0.5cm}
	\includegraphics[width=0.95\textwidth]{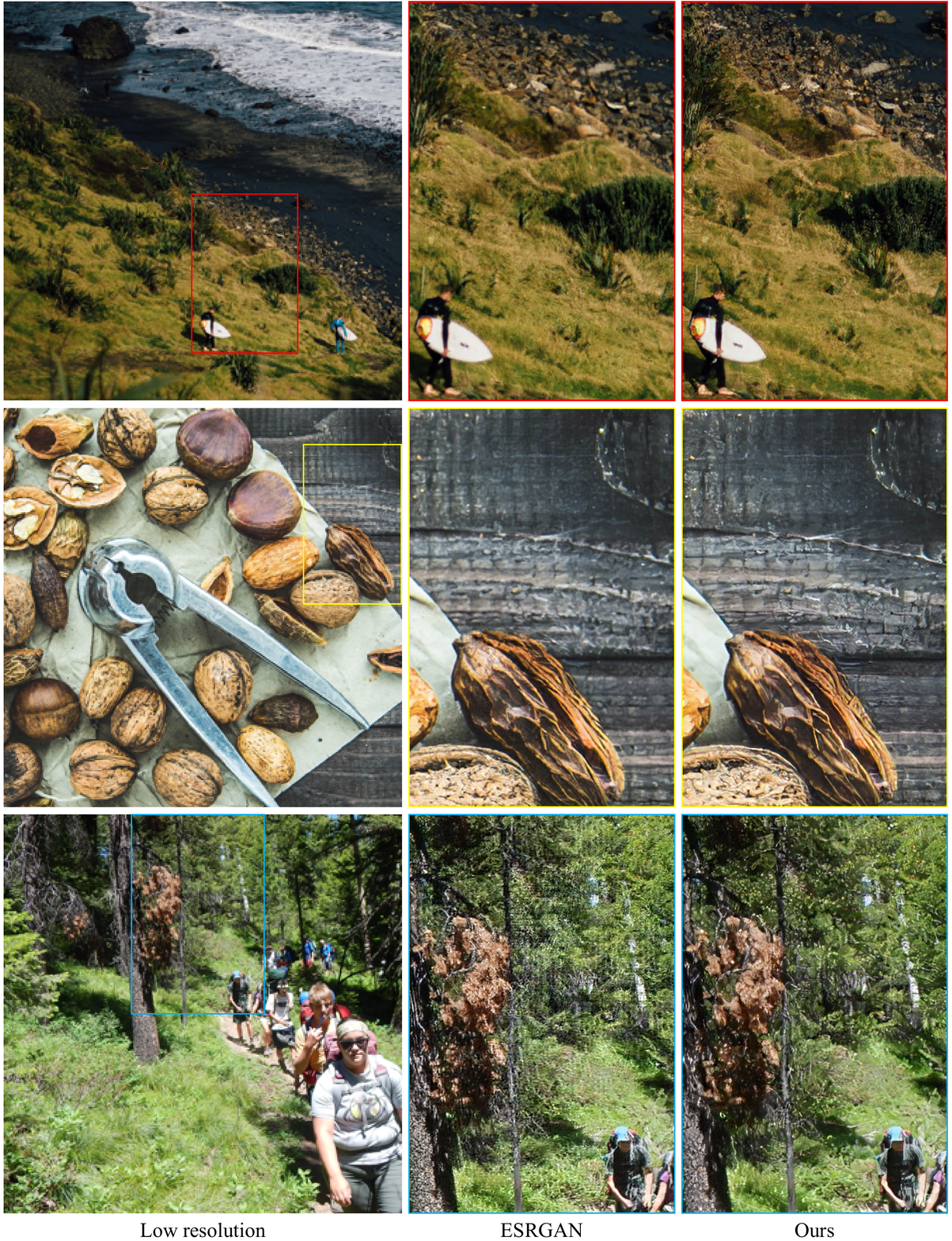}
	\vspace{-0.2cm}
	\caption{\textbf{Additional super-resolution comparisons.}}
	\label{fig:sr}
\end{figure*}

\newpage

\begin{table*}[thp]
	\begin{center}
		\begin{tabular}{ c  c }
			Generator & Critic \\  
			\begin{tabular}{ c }
				\hline \hline
				\\
				$z \in \mathbb{R}^{128} \sim \mathit{N}(0,I)$ \\
				dense $\Rightarrow M_g \times M_g \times 512$\\
				$3\times 3$, stride=2 deconv. BN 256 ReLU\\
				$3\times 3$, stride=2 deconv. BN 128 ReLU\\
				$3\times 3$, stride=2 deconv. BN 64 ReLU\\
				$3\times 3$, stride=1 conv. 3 Tanh\\ 
				\\ \\ \\
			\end{tabular} &  
			\begin{tabular}{ c } 
				\hline \hline
				\\
				RGB image $x \in \mathbb{R}^{M\times M \times 3}$\\
				$3\times 3$, stride=1 conv. 64 lReLU\\
				$3\times 3$, stride=2 conv. 64 lReLU\\
				$3\times 3$, stride=1 conv. 128 lReLU\\
				$3\times 3$, stride=2 conv. 128 lReLU\\
				$3\times 3$, stride=1 conv. 256 lReLU\\
				$3\times 3$, stride=2 conv. 256 lReLU\\
				$3\times 3$, stride=1 conv. 512 lReLU\\
				dense $\Rightarrow 1$\\
			\end{tabular}
			\\ & \\
		\end{tabular}
		\caption{\textbf{The standard architecture used for image generation.} We use a standard CNN model for our unconditional image generation experiments on CIFAR-10 and STL-10. The slopes of all leaky-ReLU (lReLU) functions are set to $0.1$. For the generator, we use  $M_g=4$ for CIFAR-10 and $M_g=6$ for STL-10. For the critic, we use $M=32$ for CIFAR-10 and $M=48$ for STL-10.}
		\label{tab:arch_gen_standard}
	\end{center}
\end{table*}

\begin{table*}[thp]
	\begin{center}
		\begin{tabular}{ c  c }
			Generator & Critic \\  
			\begin{tabular}{ c }
				\hline \hline
				\\
				$z \in \mathbb{R}^{128} \sim \mathit{N}(0,I)$ \\
				dense $\Rightarrow 4 \times 4 \times 256$\\
				ResBlock, up-sample, 256\\
				ResBlock, up-sample, 256\\
				ResBlock, up-sample, 256\\
				BN, ReLU, $3\times 3$ conv., 3 Tanh\\ 
				\\ \\
			\end{tabular} &  
			\begin{tabular}{ c } 
				\hline \hline
				\\
				RGB image $x \in \mathbb{R}^{32\times 32 \times 3}$\\
				ResBlock, down-sample, 128\\
				ResBlock, down-sample, 128\\
				ResBlock, 128\\
				ResBlock, 128\\
				ReLU\\
				Global avg. pooling\\
				dense $\Rightarrow 1$\\
			\end{tabular}
			\\ & \\
		\end{tabular}
		\caption{\textbf{The ResNet architecture used for image generation on CIFAR-10.} The ResBlock comprises a sequences of batch-norm, ReLU, convolution, batch-norm, ReLU and convolution layers. In the critic, batch-norm layers are removed from the ResBlock.}
		\label{tab:arch_resnet_cifar}
	\end{center}
\end{table*}

\begin{table*}[thp]
	\begin{center}
		\begin{tabular}{ c  c }
			Generator & Critic \\  
			\begin{tabular}{ c }
				\hline \hline
				\\
				$z \in \mathbb{R}^{128} \sim \mathit{N}(0,I)$ \\
				dense $\Rightarrow 6 \times 6 \times 512$\\
				ResBlock, up-sample, 256\\
				ResBlock, up-sample, 128\\
				ResBlock, up-sample, 64\\
				BN, ReLU, $3\times 3$ conv., 3 Tanh\\ 
				\\ \\
			\end{tabular} &  
			\begin{tabular}{ c } 
				\hline \hline
				\\
				RGB image $x \in \mathbb{R}^{48\times 48 \times 3}$\\
				ResBlock, down-sample, 64\\
				ResBlock, down-sample, 128\\
				ResBlock, down-sample, 256\\
				ResBlock, down-sample, 512\\
				ReLU\\
				Global avg. pooling\\
				dense $\Rightarrow 1$\\
			\end{tabular}
			\\ & \\
		\end{tabular}
		\caption{\textbf{The ResNet architecture used for image generation on STL-10.} The ResBlock comprises a sequences of batch-norm, ReLU, convolution, batch-norm, ReLU and convolution layers. In the critic, batch-norm layers are removed from the ResBlock. In comparison to \cite{spectral}, we remove the discriminator's last ResBlock.}
		\label{tab:arch_resnet_stl}
	\end{center}
\end{table*}

\begin{table*}[thp]
	\begin{center}
		\begin{tabular}{ c  c }
			Generator & Critic \\  
			\begin{tabular}{ c }
				\hline \hline
				\\
				RGB image $x \in \mathbb{R}^{128\times 128 \times 3}$\\
				$7\times 7$, stride=1 conv. IN 64 ReLU\\
				$4\times 4$, stride=2 conv. IN 128 ReLU\\
				$4\times 4$, stride=2 conv. IN 256 ReLU\\
				ResBlock, 256\\
				ResBlock, 256\\
				ResBlock, 256\\
				ResBlock, 256\\
				ResBlock, 256\\
				ResBlock, 256\\
				$4\times 4$, stride=2 deconv. IN 128 ReLU\\
				$4\times 4$, stride=2 deconv. IN 64 ReLU\\
				$7\times 7$ conv., 3 Tanh\\ 
			\end{tabular} &  
			\begin{tabular}{ c } 
				\hline \hline
				\\
				RGB image $x \in \mathbb{R}^{128\times 128 \times 3}$\\
				$3\times 3$, stride=2 conv. 64 lReLU\\
				$3\times 3$, stride=2 conv. 128 lReLU\\
				$3\times 3$, stride=2 conv. 256 lReLU\\
				$3\times 3$, stride=2 conv. 512 lReLU\\
				$3\times 3$, stride=2 conv. 1024 lReLU\\
				$3\times 3$, stride=2 conv. 2048 lReLU\\
				$3\times 3$, conv. 1 | $2\times 2$, conv. $n_d$\\
				\\ \\ \\ \\ \\
			\end{tabular}
			\\ & \\
		\end{tabular}
		\caption{\textbf{Image-to-image translation architectures.} For the generator, we perform instance normalization (IN) in all layers except the last output layer. The ResBlock comprises a sequences of convolution, instance-norm, ReLU, convolution and instance-norm layers. In the critic, we use leaky ReLU with slope $0.01$. Here, $n_d$ represents the number of domains, as the critic outputs a 1-hot classification vector of length $n_d$.}
		\label{tab:arch_resnet_stargan}
	\end{center}
\end{table*}

\begin{table*}[thp]
	\begin{center}
		\begin{tabular}{ c  c }
			Generator & Critic \\  
			\begin{tabular}{ c }
				\hline \hline
				\\
				RGB image $x \in \mathbb{R}^{\frac{N}{4}\times \frac{N}{4} \times 3}$\\
				$3\times 3$, stride=1 conv. 64 lReLU\\
				ResBlock, 64\\
				ResBlock, 64\\
				$\vdots$ \\
				ResBlock, 64\\
				$3\times 3$, stride=1 conv. 256, PS, lReLU\\
				$3\times 3$, stride=1 conv. 256, PS, lReLU\\
				$3\times 3$, stride=1 conv. 64, lReLU\\
				$3\times 3$, stride=1 conv. 3, +up-sample(x)\\
			\end{tabular} &  
			\begin{tabular}{ c } 
				\hline \hline
				\\
				RGB image $x \in \mathbb{R}^{128\times 128 \times 3}$\\
				$3\times 3$, stride=1 conv. 64 lReLU\\
				$3\times 3$, stride=2 conv. 64 lReLU\\
				$3\times 3$, stride=1 conv. 128 lReLU\\
				$3\times 3$, stride=2 conv. 128 lReLU\\
				$3\times 3$, stride=1 conv. 256 lReLU\\
				$3\times 3$, stride=2 conv. 256 lReLU\\
				$3\times 3$, stride=1 conv. 512 lReLU\\
				$3\times 3$, stride=2 conv. 512 lReLU\\
				dense $\Rightarrow 100$, lReLU\\
				dense $\Rightarrow 1$\\
			\end{tabular}
			\\ & \\
		\end{tabular}
		\caption{\textbf{Single image super-resolution architectures.} We use leaky ReLU with slope $0.1$, and pixelshuffle (PS) for the up-sampling operation. The ResBlock comprises a sequences of convolution, ReLU, and convolution layers. The generator consists of $16$ residual blocks as in \cite{srgan}.}
		\label{tab:arch_sr}
	\end{center}
\end{table*}

\end{document}